\documentclass[acmtog,timestamp]{acmart} 

\usepackage{booktabs}

\setcitestyle{nosort,square} 

\acmJournal{TOG}
\usepackage{hyperref}
\usepackage{amsmath}
\usepackage{graphicx}
\usepackage{tikz}
\usepackage{comment}
\usepackage{kbordermatrix}
\usepackage{multirow,bigdelim}

\usepackage{adjustbox}
\usepackage{float}
\usepackage{dblfloatfix} 
\usepackage{array}
\usepackage{lipsum}
\usepackage{wrapfig}

\newcolumntype{C}[1]{>{\centering\let\newline\\\arraybackslash\hspace{0pt}}m{#1}}

\newif\ifdraft
\drafttrue
\draftfalse

\ifdraft
\newcommand{\nfc}[1]{{\color{cyan}\textbf{NF:} #1}}
\newcommand{\dcc}[1]{{\color{red}\textbf{DC:} #1}}
\newcommand{\rhc}[1]{{\color{orange}\textbf{RH:} #1}}
\newcommand{\rgc}[1]{{\color{olive}\textbf{RG:} #1}}
\newcommand{\shflc}[1]{{\color{magenta}\textbf{SF:} #1}}
\newcommand{\ahc}[1]{{\color{purple}\textbf{AH:} #1}}

\newcommand{\nf}[1]{{\color{blue}#1}}
\newcommand{\dc}[1]{{\color{red}#1}}
\newcommand{\rh}[1]{{\color{orange}#1}}
\newcommand{\rg}[1]{{\color{olive}#1}}

\else
\newcommand{\nfc}[1]{}
\newcommand{\dcc}[1]{}
\newcommand{\rhc}[1]{}
\newcommand{\rgc}[1]{}
\newcommand{\shflc}[1]{}
\newcommand{\ahc}[1]{}
\newcommand{\nf}[1]{{\color{black}#1}}
\newcommand{\dc}[1]{{\color{black}#1}}
\newcommand{\rh}[1]{{\color{black}#1}}
\newcommand{\rg}[1]{{\color{black}#1}}

\fi

\usepackage[shortlabels]{enumitem}
                    \setlist[enumerate, 1]{1\textsuperscript{o}}

\newcommand{\ourmethod}{MeshCNN}

\newcommand{\onering}{$1$-ring}
\begin{document}

\title{MeshCNN: A Network with an Edge}

\author{Rana Hanocka}
\affiliation{\institution{Tel Aviv University}}

\author{Amir Hertz}
\affiliation{\institution{Tel Aviv University}}

\author{Noa Fish}
\affiliation{\institution{Tel Aviv University}}

\author{Raja Giryes}
\affiliation{\institution{Tel Aviv University}}

\author{Shachar Fleishman}
\affiliation{\institution{Amazon}}

\author{Daniel Cohen-Or}
\affiliation{\institution{Tel Aviv University}}

\begin{abstract}
%


%


Polygonal meshes provide an efficient representation for 3D shapes. They explicitly capture both shape surface and topology, and leverage non-uniformity to represent large flat regions as well as sharp, intricate features. This non-uniformity and irregularity, however, inhibits mesh analysis efforts using neural networks that combine convolution and pooling operations.
In this paper, we utilize the unique properties of the mesh for a direct analysis of 3D shapes using \emph{\ourmethod{}}, a convolutional neural network designed specifically for triangular meshes. Analogous to classic CNNs, \ourmethod{} combines specialized convolution and pooling layers that operate on the mesh edges, by leveraging their intrinsic geodesic connections. Convolutions are applied on edges and the four edges of their incident triangles, and pooling is applied via an edge collapse operation that retains surface topology, thereby, generating new mesh connectivity for the subsequent convolutions. 
\ourmethod{} learns which edges to collapse, thus forming a task-driven process where the network 
exposes and expands the important features while discarding the redundant
ones.
We demonstrate the effectiveness of our task-driven pooling on various learning tasks applied to 3D meshes.

\end{abstract}

\maketitle

\section{Introduction}

Three dimensional shapes are front and center in the field of computer graphics, but also a major commodity in related fields such as computer vision and computational geometry.
Shapes around us, and in particular those describing natural entities, are commonly composed of continuous surfaces. 

For computational reasons, and to facilitate data processing, various discrete approximations for 3D shapes have been suggested and utilized to represent shapes in an array of applications. A favorite of many, the polygonal mesh representation, or mesh, for short, approximates surfaces via a set of 2D polygons in 3D space \cite{ botsch2010polygon}. The mesh provides an efficient, non-uniform representation of the shape. On the one hand, only a small number of polygons are required to capture large, simple, surfaces. On the other hand, representation flexibility supports a higher resolution where needed, allowing a faithful reconstruction, or portrayal, of salient shape features that are often geometrically intricate. Another lucrative characteristic of the mesh is its native provision of connectivity information. This forms a \rh{comprehensive}
representation of the underlying surface.

These advantages are apparent in comparison to another popular option: the point cloud representation. Despite its simplicity and direct relation to common data acquisition techniques (scanning), the point cloud representation falls short when a higher quality and preservation of sharp shape features are required.

\begin{figure}[t!]
\newcommand{\vfig}{8}
\begin{center}
\begin{tabular}{c}
\adjincludegraphics[width=\vfig cm,trim={{0.05\width} {.13\height} {.08\width} {.13\height}},clip]{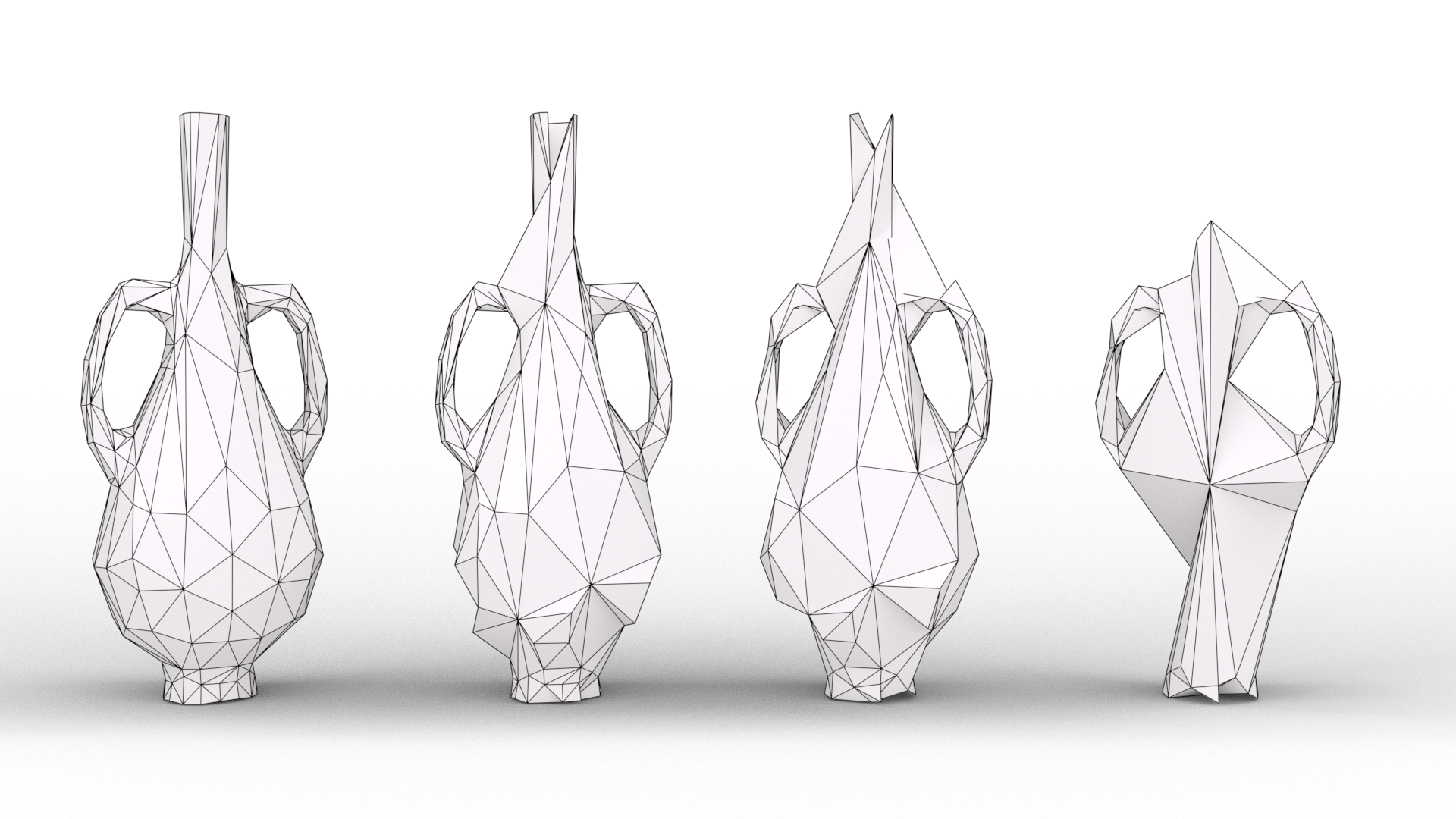} \\
\adjincludegraphics[width=\vfig cm,trim={{0.05\width} {.13\height} {.08\width} {.13\height}},clip]{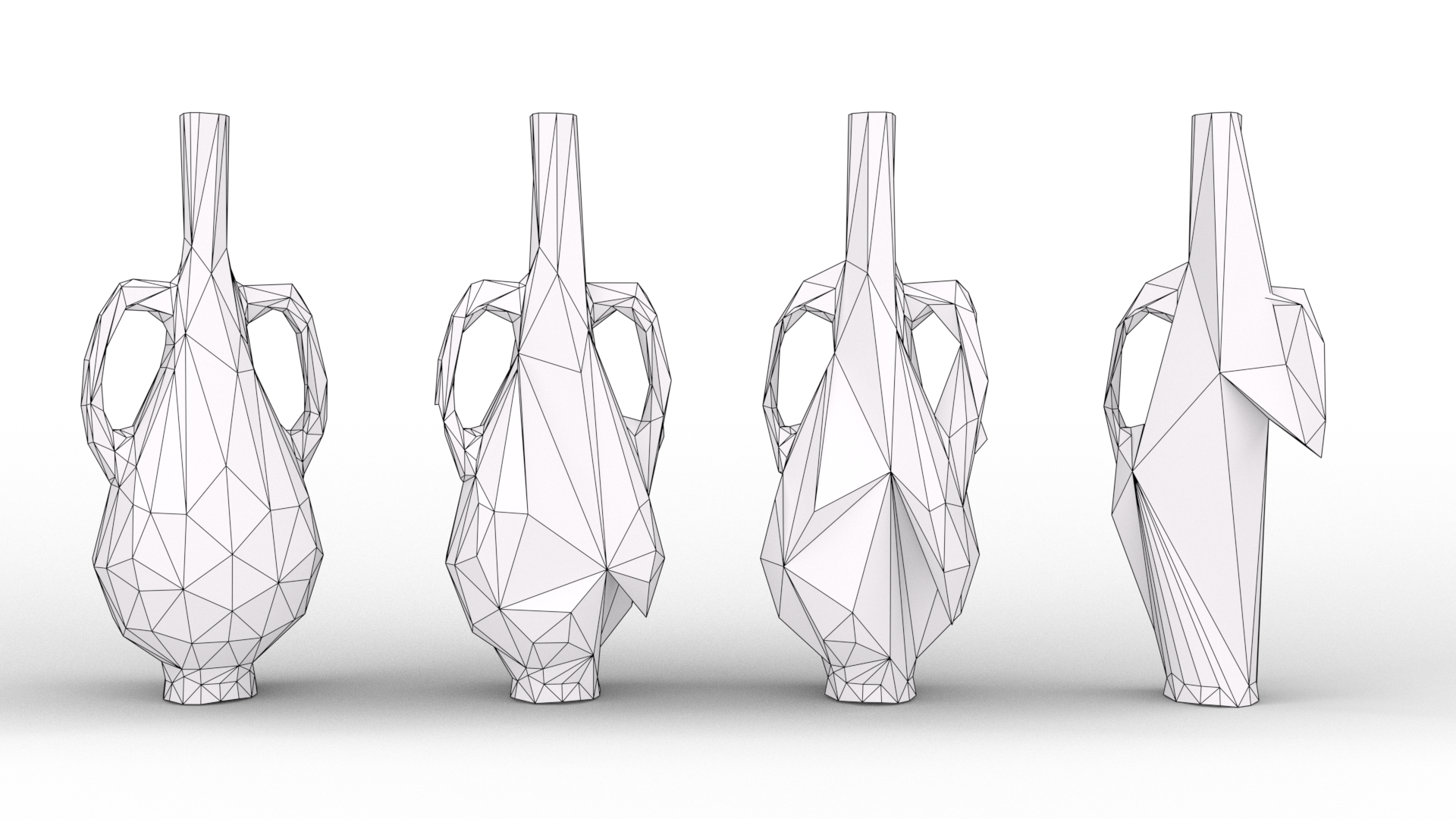} \\
\end{tabular}
\end{center}
\caption{Mesh pooling operates on irregular structures and adapts spatially to the task. Unlike geometric simplification (removes edges with a minimal geometric distortion), mesh pooling delegates which edges to collapse to the network. Top row: \ourmethod{} trained to classify whether a vase has a \textit{handle}, bottom row: trained on whether there is a \textit{neck} (top-piece). } 
\label{fig:teaser}
\end{figure}

\begin{figure*}[!ht]
\includegraphics[width=15cm]{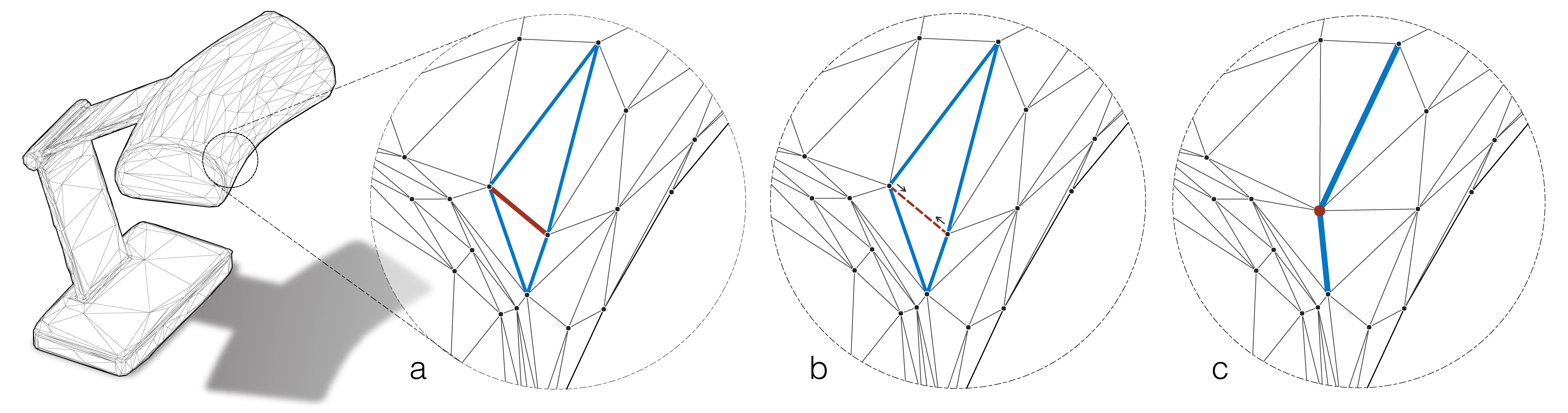}
\caption{
(a) Features are computed on the edges by applying convolutions with neighborhoods made up of the four edges of the two incident triangles of an edge. The four blue edges are incident to the red edge. The pooling step is shown in (b) and (c). In (b), the red edge is collapsing to a point, and the four incident (blue) edges merge into the two (blue) edges in (c). Note that in a single edge collapse step, five edges are converted into two.}
\label{fig:coco}
\end{figure*}

In recent years, using convolutional neural networks (CNNs) on images has demonstrated outstanding performance on a variety of tasks such as classification and semantic segmentation \shortcite{sermanet2013overfeat,vgg,chen2018deeplab}.
The recipe for their success features a combination of convolution, non-linearity and pooling layers, resulting in a framework \rh{that is invariant (or \textit{robust}) to irrelevant variations of the input}~\cite{lecun2012learning, krizhevsky2012imagenet}.
However, since images are represented on a regular grid of discrete values, extending CNNs to work on irregular structures is nontrivial.

Initial approaches bypassed adapting CNNs to irregular data by using regular representations: mapping 3D shapes to multiple 2D projections~\cite{Su15Multi} or 3D voxel grids \cite{Wu15Shapenet}.
While these approaches benefit from directly using well understood image CNN operators, their indirect representation requires prohibitively large amounts of memory with wasteful or redundant CNN computations (\emph{e.g.,} convolutions on unoccupied voxels). 

More efficient approaches directly apply CNNs on the irregular and sparse point cloud representation~\cite{qi2017pointnet}. While these approaches benefit from a compact input representation, they are inherently oblivious to the local surface. Moreover, the notion of neighborhoods and connectivity is ill-defined, making the application of convolution and pooling operations nontrivial. This ambiguity has resulted in a wave of works geared towards overcoming this challenge~\cite{monti2017geometric,dgcnn,Li18PointCNN,Yi17SyncSpecCNN}.

Aiming to tap into the natural potential of the native mesh representation, we present \emph{\ourmethod{}}: a neural network akin to the well-known CNN, but designed specifically for meshes.
\ourmethod{} operates directly on irregular triangular meshes, performing convolution and pooling operations designed in harmony with the unique mesh properties.
In \ourmethod{}, the edges of a mesh are analogous to pixels in an image, since they are the basic building blocks which all operations are applied on.
\nf{We choose to work with edges since every edge is incident to exactly two faces (triangles), which defines a natural fixed-sized convolutional neighborhood of four edges (see Figure \ref{fig:coco}).}
\rh{We utilize the consistent face normal order to apply a
symmetric convolution operation, which learns edge features that
are invariant to transformations in rotation, scale and translation.}

A key feature of \ourmethod{} is the unique pooling operation, \emph{mesh pooling}, which operates on irregular structures and spatially adapts to the task.
In CNNs, pooling downsamples the number of features in the network, thereby learning to eliminate less informative features.
Since features are on the edges, an intuitive approach for down-sampling is to use the well-known mesh simplification technique \emph{edge collapse} \cite{hoppe1997view}.
However, unlike conventional edge collapse, which removes edges that introduce a minimal geometric distortion, mesh pooling delegates the choice of which edges to collapse to the network \rg{in a task-specific manner.
The purged edges are the ones whose features contribute the least to the used objective}
(see examples in \rg{figures} \ref{fig:teaser} and \ref{fig:pool_cubes}). 

To increase flexibility and support a variety of available data, each pooling layer simplifies the mesh to a predetermined constant number of edges. 
\rg{Moreover, while it produces any prescribed output edge count, 
\ourmethod{} is agnostic to the input mesh size and capable of handling different \dc{triangulations}.}
We show, as demonstrated in Figure~\ref{fig:teaser}, that intermediate computational \dc{pooling} steps, as well as the final output are semantically interpretable.
To illustrate the aptitude of our method, we perform a variety of experiments on shape classification and segmentation tasks and demonstrate superior results to state-of-the-art approaches on common datasets and on highly non-uniform meshes.
\section{Related Works}
Many of the operators that we present or use in our work are based on classic mesh processing techniques \cite{hoppe1999new,Rusinkiewicz:2000:QMP:344779.344940, botsch2010polygon, kalogerakis2010learning}, or more specifically, mesh simplification techniques \cite{hoppe1993mesh,garland1997surface, hoppe1997view}.
In particular, we use the edge-collapse technique \cite{hoppe1997view} for our task-driven pooling operator.
While classic mesh simplification techniques aim to reduce the number of mesh elements with minimal geometric distortion~\cite{tarini2010practical, gao2017robust}, in this work we use the mesh simplification technique to reduce the resolution of the feature maps within the context of a neural network. In the following, we revisit relevant work on 3D data analysis using neural networks, organized according to input representation type.

{\bf Multi-view 2D projections.} Leveraging existing techniques and architectures from the 2D domain is made possible by representing 3D shapes through their 2D projections from various viewpoints. These sets of rendered images serve as input to subsequent processing by standard CNN models. Su et al. \shortcite{Su15Multi} were the first to apply a multi-view CNN for the task of shape classification, however, this approach (\emph{as is}) cannot perform semantic-segmentation.  
Later,~\cite{kalogerakis20173d} suggested a more comprehensive multi-view framework for shape segmentation: generating image-level segmentation maps per view and then solving for label consistency using CRF (trained end-to-end).
Qi et al. \shortcite{Qi16Volumetric} explored both view-based and volumetric approaches, and observed the superiority of the first \rg{compared to the methods available at that time}.

{\bf Volumetric.} Transforming a 3D shape into a binary voxel form provides a grid-based representation that is analogous to the 2D grid of an image. As such, operations that are applied on 2D grids can be extended to 3D grids in a straight-forward manner, thus allowing a natural transference of common image-based approaches to the shape domain. Wu et al. \shortcite{Wu15Shapenet} pioneered this concept, and presented a CNN that processes voxelized shapes for classification and completion. Following that, Brock et al. \shortcite{Brock16Generative} tackled shape reconstruction using a voxel-based variational autoencoder, and \cite{Tchapmi17SEGCloud} combined trilinear interpolation and Conditional Random Fields (CRF) with a volumetric network to promote semantic shape segmentation. Hanocka et al. \shortcite{hanocka2018alignet} used volumetric shape representations to train a network to regress grid-based warp fields for shape alignment, and applied the estimated deformation on the original mesh. 

Despite their alluring simplicity, volumetric representations are computationally demanding, requiring significant memory usage. To alleviate this, several acceleration strategies have been proposed, where sparsity of shape occupancy within the volume is exploited for representation reduction \cite{Li16FPNN,Riegler17OctNet,Wang17OCNN,Graham17Semantic}

{\bf Graph.} A common generalization of grid-based representations that allows non-regularity, is the graph structure. To support graph-based data analysis, considerable focus has been directed toward the application of neural networks to popular tasks involving data represented in graph form, mainly, social networks, sensor networks in communication, or genetic data. One approach advocates for the processing of the Laplacian of the graph representation \cite{Bruna14Spectral, Henaff15Deep, defferrard2016convolutional, kostrikov2018surface}, and thus operates in the spectral domain.
Another approach opts to process the graph directly by extracting locally connected regions and transforming them into a canonical form to be processed by a neural network \cite{Niepert16Learning}. Atwood et al. \shortcite{Atwood16Diffusion} proposed diffusion-convolution, where diffusion is applied on each node to determine its local neighborhood. 
Monti et al. \shortcite{monti2017geometric} parameterize the surface into local patches using the graph spatial domain.
Xu et al.~\shortcite{xu2017directionally} use directional convolutions on surface patches for the task of semantic segmentation.
Yi et al.~\shortcite{Yi17SyncSpecCNN} use graph convolutions in the spectral domain on the task of 3D segmentation.
Kostrikov et al.~\shortcite{kostrikov2018surface} use a laplacian surface network for developing a generative model for 3D shapes.
Such et al. \shortcite{Such17Robust} introduced the concept of vertex filtering on graphs, but did not incorporate pooling operations for feature aggregation. \rh{These methods commonly operate on the \emph{vertices} of a graph.}

{\bf Manifold.} The pioneering work of Masci et al. \shortcite{Masci15Geodesic} introduced deep learning of local features on meshes \rg{(intrinsic mesh descriptors similar to the ones in \cite{Kokkinos12Intrinsic})}, and has shown how to use these learned features for correspondence and retrieval. Specifically, they demonstrate how to make the convolution operations intrinsic to the mesh.

Often, local patches on a manifold shape are approximately Euclidean. This characteristic can be exploited for manifold shape analysis using standard CNNs, by parameterizing the 3D manifold to 2D~\cite{Henaff15Deep, boscaini2016learning, sinha2016deep, maron2017convolutional,gwcnn}. 
Boscaini et al.~\shortcite{boscaini2015learning} use vertex-frequency analysis to learn a local intrinsic 3D shape descriptor. 
Another approach uses a toric topology to define the convolutions on the shape graph \cite{Haim2018Surface, maron2017convolutional}. Poulenard et al. \shortcite{Poulenard} define a new convolutional layer that allows propagating geodesic information throughout the layers of the network.

Verma et al. \shortcite{Verma2018FeaStNetFG} proposed a graph neural network in which the neighborhood of each vertex for the convolution operation is not predefined but rather calculated dynamically based on its features. 
Tatarchenko et al. \shortcite{Tat18} introduced tangent convolution, where a small neighborhood around each point is used to reconstruct the local function upon which convolution is applied. Unlike previous works, they incorporated pooling operations by subsampling on a regular 3D grid. 
Some generative models have been also proposed. Litany et. al. \shortcite{Litany2018DeformableSC} introduce an autoencoder that performs shape completion. Ranjan et. al. \shortcite{Ranjan18Generating} demonstrate how 3D faces \rh{can} be generated by mesh autoencoders.

See~\cite{Bronstein17Geometric} for a comprehensive survey on geometric deep learning. The uniqueness of our approach compared to the previous ones is that our network operations are \rh{specifically} designed to adapt to the mesh structure. In particular, we learn a unique pooling operator that selects which areas to simplify, based on the target task.

To the best of our knowledge this is the first work that proposes (i) a convolution on the edges of a mesh and (ii) a learned mesh pooling operation that adapts to the task at hand. 
In \cite{Ranjan18Generating}, a fixed pooling operation has been proposed for meshes autoencoders. Learned pooling operations have been proposed in the context of graph neural networks \cite{Ying18Hierarchical, Cangea18Hierarchical}. Yet, these operations do not take into account the unique triangular mesh properties.

A convolution that extracts edge features has been proposed with the dual graph convolution models \cite{Monti2018DualPrimalGC} that extend the graph attention networks \cite{Velickovic2018graph}. \rh{Yet, the attention and optimization} mechanisms \rh{used in their work are} very different than ours; in this work we define operators for meshes, which \rg{exploit} 
their unique structure and properties. This allows us to define a symmetric convolution that leads \rg{to} invariant network features.

{\bf Point clouds.} Arguably the simplest of all representations, the point cloud representation provides a no-frills approximation for an underlying 3D shape. The close relationship to data acquisition, and ease of conversion from other representations, make the point cloud form a classic candidate for data analysis. Accordingly, recent effort has focused on developing techniques for point cloud shape analysis using neural networks. PointNet \cite{qi2017pointnet} proposes to use $1x1$ convolutions followed by global max pooling for order invariance. In its followup work, PointNet++ \cite{Qi17PointNetpp}, points are partitioned to capture local structures better. Wang et al. \shortcite{wang2018dynamic} take into account local point neighborhood information, and perform dynamic updates driven by similarity calculation between points based on distance in feature space. 
While most point-based approaches focus on global or mid-level attributes, ~\cite{GuerreroEtAl:PCPNet:EG:2018} proposed a network to estimate local shape properties, \emph{e.g.,} normal and curvature from raw point clouds,
while~\cite{williams2018deep} learn a geometric prior for surface reconstruction from point clouds.
Atzmon et al \shortcite{atzmon2018point} defines an efficient convolution operator on point clouds by mapping the point clouds functions into volumetric functions. This makes the method invariant to the order of points and robust to some deformations in the data.
Recently, Li et al. \shortcite{Li18PointCNN} presented PointCNN, which extends the notion of convolution from a local grid to an $\chi$-convolution on points residing in its Euclidean neighborhood.

\rh{In this work, unlike previous work, \rg{we rely on mesh edges to provide non-uniform, geodesic neighborhood information} and have a consistent number of conv-neighbors.} \dc{Invariant} feature computation is performed on the edges, while leveraging mesh decimation techniques, such as edge collapse, in a manner that adheres to shape geometry and topology.

\section{Overview: Applying CNN on Meshes}
\label{sec:overview}
\begin{figure}[t!]
\begin{tabular}{cc}
\adjincludegraphics[width=3.5 cm,trim={{0.04\width} {.05\height} {.05\width} {.05\height}},clip]{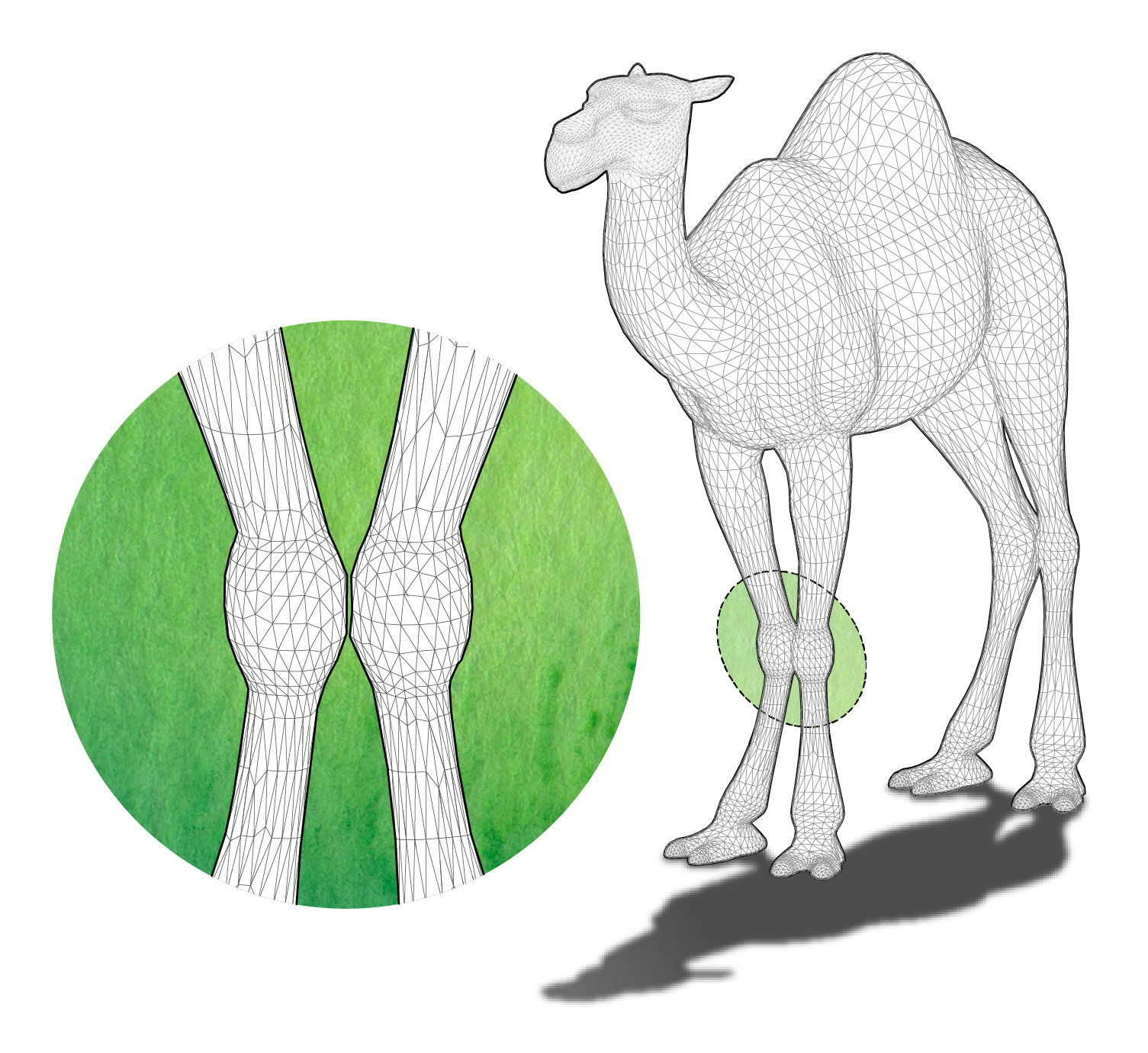} &
\adjincludegraphics[width=3.5 cm,trim={{0.04\width} {0.0\height} {0.05\width} {0.07\height}},clip]{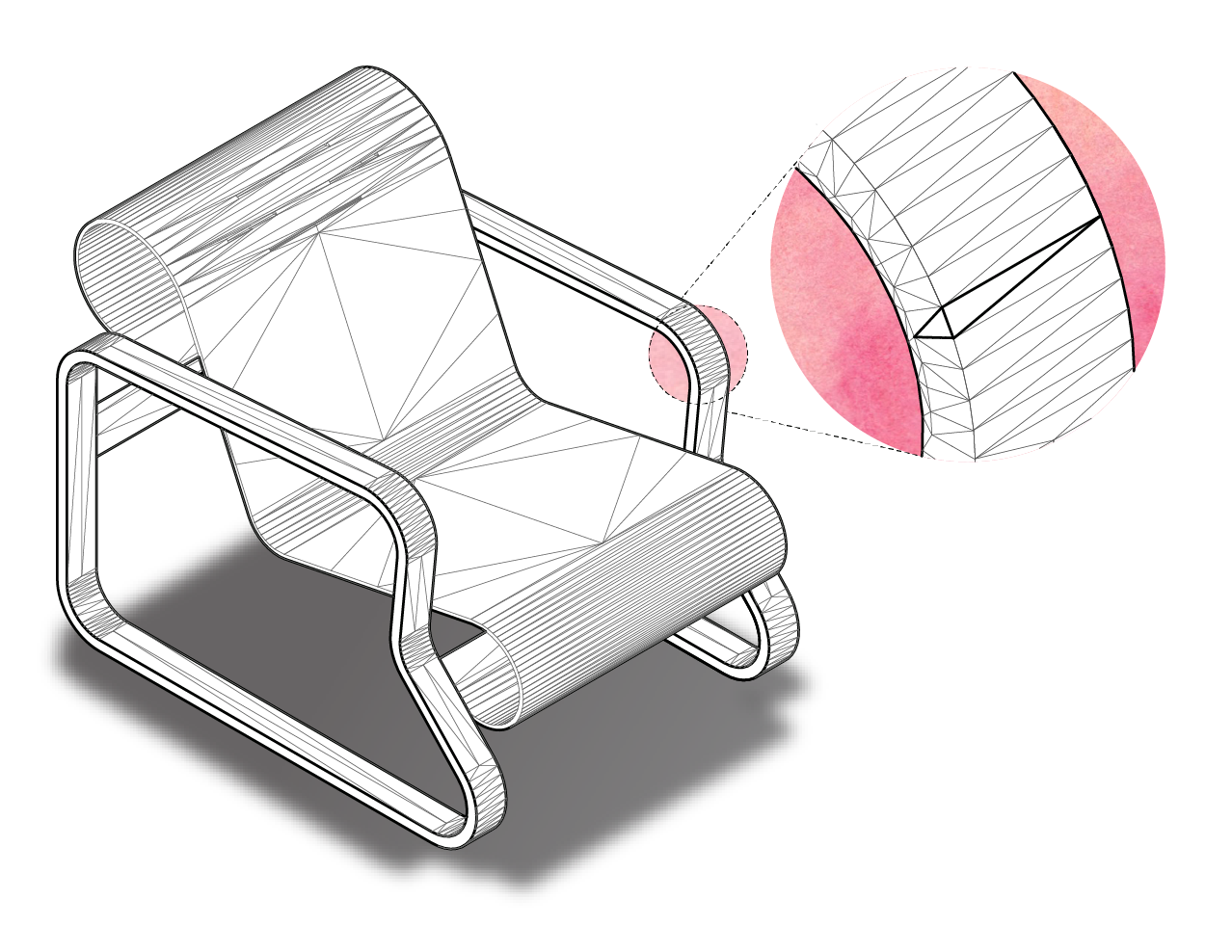} \\
\end{tabular}
\caption{
\nf{Polygonal mesh advantages. \textit{Left}: accurate portrayal of shape structure. The mesh, unlike the point cloud, can easily convey the distinct identities of the camel joints (zoom-in) through geodesic separation, despite their proximity in Euclidean space. \textit{Right}: adaptive non-uniform representation. Large flat regions can be represented by a small number of large polygons, with detailed 
regions represented by a larger number of small polygons.} }
\label{fig:euc_geod}
\end{figure}

The most fundamental and commonly used 3D data representation in computer graphics is the non-uniform polygonal mesh; large flat regions use a small number of large polygons, while detailed regions use a larger number of polygons.
A mesh explicitly represents the topology of a surface: faithfully describing intricate structures while disambiguating proximity from nearby surfaces (see Figure \ref{fig:euc_geod}). 
%

Realizing our goal to apply the CNN paradigm directly onto triangular meshes, necessitates an analogous definition and implementation of the standard \rg{building blocks} of CNN: the convolution and pooling layers.
As opposed to images which are represented on a regular grid of discrete values, the key challenge in mesh analysis is its inherent irregularity and non-uniformity. 
In \rh{this} work, we aim to exploit these challenging and unique properties, rather than bypassing them. Accordingly, we design our network to deliberately apply convolution and pooling operations directly on the constructs of the mesh, and avoid conversion to a regular and uniform representation.

\begin{wrapfigure}{r}{0.09\textwidth}
\includegraphics[width=0.09\textwidth]{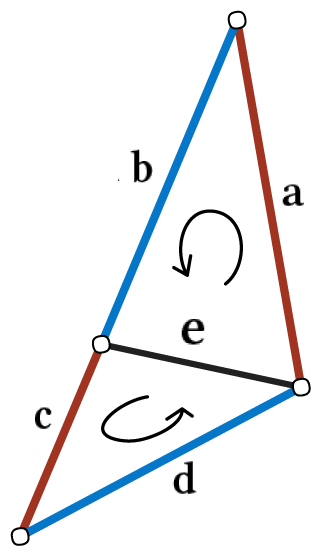}
\caption{ }
\label{fig:triangle_neigh}
\vspace{-10pt}
\end{wrapfigure}
\textbf{Invariant convolutions.} In our setting, all shapes are assumed to be represented as manifold meshes, possibly with boundary edges. Such an assumption guarantees that each edge is incident to two faces (triangles) at most, and is therefore adjacent to either two or four other edges.
\nf{The vertices of a face are ordered counter-clockwise, defining two possible orderings for the four neighbors of every edge. For example, see Figure~\ref{fig:triangle_neigh}, where the \onering{} neighbors of $e$ can be ordered as $(a,b,c,d)$ or $(c,d,a,b)$, depending on which face defines the first neighbor. This ambiguates the convolutional receptive field, hindering the formation of invariant features.}

We take two actions to address this issue and guarantee invariance to similarity transformations (rotation, translation and scale) within our network. First, we carefully design the input descriptor of an edge to contain only relative geometric features that are inherently invariant to similarity transformations. Second, 
\rh{we aggregate the four \onering{} edges into two pairs of edges which have an ambiguity (\emph{e.g.,} $a$ and $c$, and $b$ and $d$), and generate new features by applying simple symmetric functions on each pair (\emph{e.g.,} $sum(a,c)$). The convolution is applied on the new symmetric features, thereby eliminating any order ambiguity.}

\begin{wrapfigure}{r}{0.09\textwidth}
\centering
\includegraphics[width=0.09\textwidth]{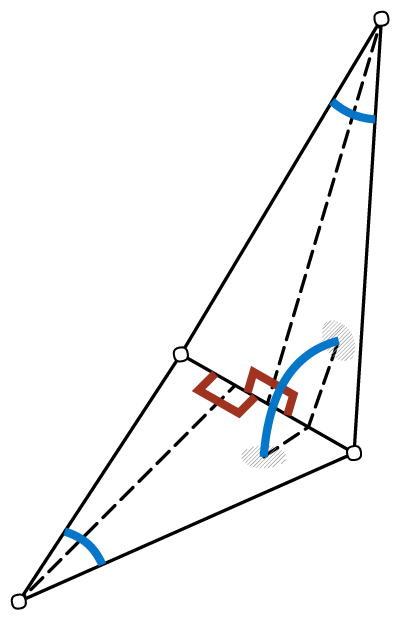}
\hspace{-10pt}
\centering
\end{wrapfigure}

\textbf{Input features.} The input edge feature is a $5$-dimensional vector for every edge: the dihedral angle, two inner angles and two edge-length ratios for each face. The edge ratio is between the length of the edge and the perpendicular (dotted) line for each adjacent face.
\nf{We sort each of the two face-based features (inner angles and edge-length ratios), thereby resolving the ordering ambiguity and guaranteeing invariance.}
Observe that these features are all \textit{relative}, making them invariant to \textit{translation}, \textit{rotation} and \textit{uniform scale}.

{\bf Global ordering.}  The global ordering of the edges is the order in which the edge data (input features) of a particular shape enters the network. This ordering has no influence during the convolutional stage since convolution is performed within local neighborhoods. By extension, fully convolutional tasks \emph{e.g.}, segmentation are unaffected by it.
\nf{For tasks that require global feature aggregation, such as classification, we follow the common practice suggested by Qi et al.~\shortcite{qi2017pointnet} in PointNet, and place a global average pooling layer that connects between the convolutional part and the fully-connected part of the network. This layer renders the initial ordering inconsequential and thus guarantees invariance to transformations.}


{\bf Pooling.} Mesh pooling is accomplished by an edge collapse process, as illustrated in Figure \ref{fig:coco} (b) and (c). In (b), the dashed edge is collapsing to a point, and, subsequently, the four incident edges (blue) merge into the two (blue) edges in (c). Note that in this edge collapse operation, five edges are transformed into two. 
The operator is prioritized by the (smallest norm) edge features, thereby allowing the network to select which parts of the mesh to simplify, and which to leave intact. This creates a task-aware process, where the network learns to determine object part importance with respect to its mission \dc{(see Figure \ref{fig:teaser})}.

A notable advantage of the nature of our simplification, is that it provides flexibility with regards to the output dimensions of the pooling layer, just before it reaches the final fully connected layer.
Pooling also contributes to robustness to initial mesh triangulation. While it does not provide equivariance to triangulation, in practice, by way of continuously collapsing edges and simplifying the mesh, we observe convergence to similar representations despite differences in initial tessellation.

\section{Method}
\label{sec:method}

A grid-based (\emph{e.g.,} image) representation conveniently provides both spatial neighbors (connectivity) and features in a single matrix. 
However, since irregular meshes do not conform to this format, we must define the features separately from the connectivity.
We accomplish this by working within the standard constructs of a mesh. 

A mesh is defined by the pair ($V$, $F$), where $V= \left \{ \mathbf{v}_1, \mathbf{v}_2 \cdots \right \}$ is the set of vertex positions in $\mathbb{R}^3$, and $F$ defines the connectivity (triplets of vertices for triangular meshes). Given ($V$, $F$), the mesh connectivity is also defined using \textit{edges} $E$, a set of pairs of vertices.

All the mesh elements $V$, $F$ and $E$ can be associated with various features (such as normals or colors). In this work, $E$ also holds a set of features. The edge features start out as a set of similarity-invariant geometric features (equivalent to RGB values in the case of an image), and develop a higher abstraction as they progress through the network layers.


\nf{In our setting, the mesh provides two attributes to the network: connectivity for convolutional neighbors and the initial geometric input features. 
The mesh vertices carry no meaning once the input features have been extracted. New vertex positions following edge collapse operations have no effect on the classification and segmentation tasks
and they are computed for visualization purposes only.}

In what follows, we expand and provide details about our mesh convolution, mesh pooling and mesh unpooling operations. 
\begin{figure}
\includegraphics[width=7cm]{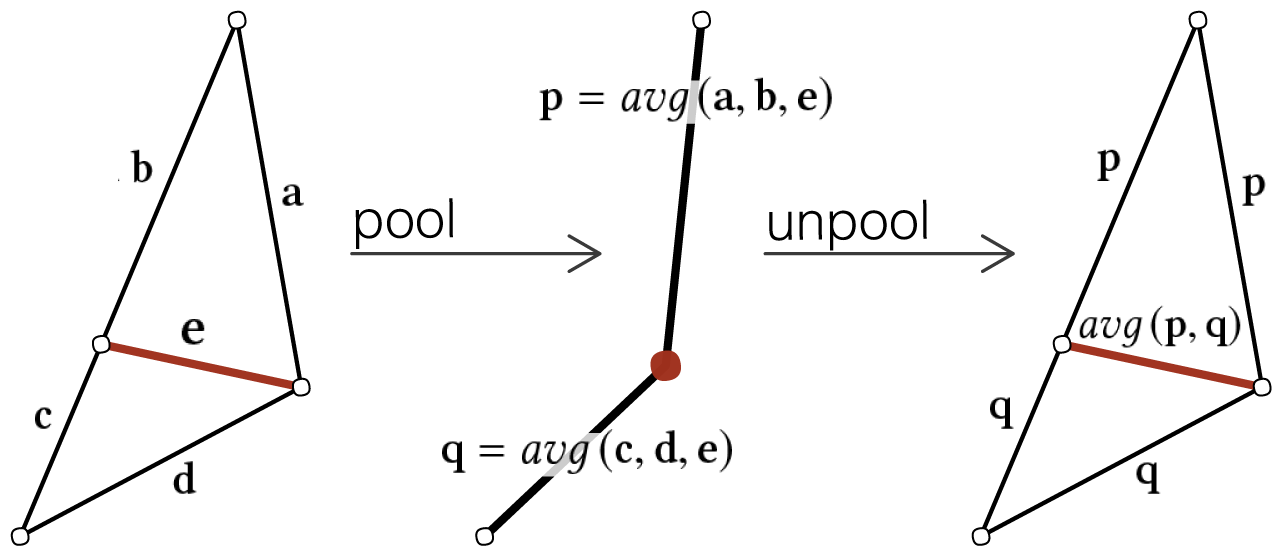} \\
\caption{
Feature aggregation during mesh pooling and unpooling.
}
\label{fig:feat_agg}
\end{figure}

\subsection{Mesh Convolution}
We define a convolution operator for edges, where the spatial support is defined using the four incident neighbors (Figure~\ref{fig:euc_geod}).
\nf{Recall that convolution is the dot product between a kernel $k$ and a neighborhood, thus the convolution for an edge feature $e$ and its four adjacent edges is:
\begin{equation}
      e \cdot k_0 +  \sum\limits_{j=1}^{4} k_j \cdot e^j,
\end{equation}
where $e^j$ is the feature of the $j^{\textrm{th}}$ convolutional neighbor of $e$. Note that as shown in Figure~\ref{fig:triangle_neigh}, the four neighbors of $e$, \emph{i.e.,} $(e^1,e^2,e^3,e^4)$, are either  $(a,b,c,d)$ or $(c,d,a,b)$, such that each filter value operates on at most two possible edges (for example $k_1$ on $a$ or $c$).}
%
%
\nf{To guarantee convolution invariance to the ordering of the input data, we apply a set of simple symmetric functions to the ambiguous pairs. This generates a new set of convolutional neighbors that are guaranteed to be invariant. In our setting, the receptive field for an edge $e$ is given by}
\begin{equation}
    (e^1,e^2,e^3,e^4) = (|a-c|, a+c, |b-d|, b+d).
\label{eq:conv_nbrs}
\end{equation}
\nf{This leads to a convolution operation that is oblivious to the initial ordering of the mesh elements, and will therefore produce the same output regardless of it.}
Recall that convolution of a multi-channel tensor with a kernel can be implemented with general matrix multiplication ($GEMM$): by expanding (or \emph{unwrapping}) the image into a column matrix (\emph{i.e.,} \emph{im2col}~\cite{jia2014learning}). Equivalently, we build an unwrapped matrix to perform the convolution operation efficiently. 

\shflc{RANA: I do not understand what you do below. I think that a figure can help. Given time limit, you may want to leave this obscure as it is, and fix for thre rebuttal.}
\rgc{I suggest using $n_c$ for channel number and $n_e$ for number of edges as now we have an ambiguity with the notation $c$ and $e$ used in Fig. 5 and other places.}
In practice, we can use highly optimized batched operators (\emph{e.g.,} \textit{conv2D}) by aggregating all edge features into a $n_c \times n_e \times 5$ feature-tensor, where $n_e$ is the number of edges, $n_c$ is the number of feature channels, and $5$ is for the edge and the convolutional neighbors (equation~\ref{eq:conv_nbrs}). This matrix is multiplied by a matrix of the weights of the convolutions using standard $GEMM$. 

Following the convolution operation, a new batched-feature-tensor is generated, where the new number of features is equal to the number of convolution kernels (just as in images). 
Note that after each pooling phase, the new connectivity will define the new convolutional neighbors for the next convolution.


\subsection{Mesh Pooling}
\label{sec:meshpooling}
We extend conventional pooling to irregular data, by identifying three core operations that together generalize the notion of pooling:
\begin{enumerate}[1)]
    \item define pooling region given adjacency
    \item merge features in each pooling region
    \item redefine adjacency for the merged features
\end{enumerate}
For pooling on regular data such as images, adjacency is inherently implied and, accordingly, the pooling region is determined directly by the chosen kernel size.
Since features in each region are merged (\emph{e.g.,} via  $avg$ or $max$) in a way that yields another uniformly spaced grid, the new adjacency is once again inherently defined. Having addressed the three general pooling operations defined above, it is evident that conventional pooling is a special case of the generalized process.

Mesh pooling is another special case of generalized pooling, where adjacency is determined by the topology. 
Unlike images, which have a natural reduction factor of, for example $4$ for $2 \times 2$ pooling, we define mesh pooling as a series of edge collapse operations, where each such edge collapse converts five edges into two. Therefore, we can control the desired \textit{resolution} of the mesh after each pooling operator, by adding a hyper-parameter which defines the number of target edges in the \textit{pooled} mesh. 
During runtime, extracting mesh adjacency information requires querying special data structures that are continually updated (see~\cite{Berg:2008} for details).

We prioritize the edge-collapse order (using a priority queue) by the magnitude of the edge features, allowing the network to select which parts of the mesh are relevant to solve the task.
\rg{This enables} the network to non-uniformly collapse certain regions which are least important to the loss. Recall that collapsing an edge which is adjacent to two faces results in a deletion of three edges (shown in Figure \ref{fig:coco}), since both faces become a single edge. Each face contains three edges: the minimum edge and two adjacent neighbors of the minimum edge (see minimum edge in red and adjacent neighbors in blue in Figure \ref{fig:coco}). Each of the features of the three edges in each face are merged into a new edge feature by taking the average over each feature channel. 

Edge collapse is prioritized according to the strength of the features of the edge, \nf{which is taken as their $\ell_2$-norm.} 
%
%
\rh{The features are aggregated as illustrated in Figure~\ref{fig:feat_agg}, where there are two merge operations, one for each of the incident triangles of the minimum edge feature $e$, resulting in two new feature vectors (denoted $p$ and $q$). The edge features in channel index $i$ for both triangles is given by}
\rh{
\begin{equation}
 p_i = avg (a_i, b_i, e_i),
 \textrm{ and, }
 q_i = avg (c_i, d_i, e_i),
\end{equation}
}
After edge collapse, \rh{the half-edge data structure is} updated for the subsequent edge collapses.

Finally, note that not every edge can be collapsed. 
An edge collapse yielding a non-manifold face is not allowed in our setting, as it violates the four convolutional neighbors assumption.
Therefore, an edge is considered invalid to collapse if it has three vertices on the intersection of its \onering{}, or if it has two boundary vertices.

\subsection{Mesh Unpooling}
Unpooling is the (partial) inverse of the pooling operation. While pooling layers reduce the resolution of the feature activations (encoding or compressing information), unpooling layers increase the resolution of the feature activations (decoding or uncompressing information). The pooling operation records the history from the merge operations (\emph{e.g.,} max locations), and uses them to expand the feature activation. Thus, unpooling does not have \emph{learnable} parameters, and it is typically combined with convolutions to recover the original resolution lost in the pooling operation. \rg{The combination with the convolution effectively makes the unpooling a learnable operation.}

Each mesh unpooling layer is paired with a mesh pooling layer, to upsample the mesh topology and the edge features. 
The unpooling layer reinstates the upsampled topology (prior to mesh pooling), by storing the connectivity prior to pooling. Note that upsampling the connectivity is a reversible operation (just as in images). 
For unpooled edge feature computation, we retain a graph which stores the adjacencies from the original edges (prior to pooling) to the new edges (after pooling). Each unpooled edge feature is then a weighted combination of the pooled edge features. \rg{The case of average unpooling is demonstrated in Figure~\ref{fig:feat_agg}}.

\section{Experiments}
\begin{figure}
\newcommand{\cfig}{8.3}
\newcommand{\tfig}{1.3}
\begin{center}
\begin{tabular}{c}
\adjincludegraphics[width=\cfig cm,trim={{0\width} {0.13\height} {0.04\width} {0.32\height}},clip]{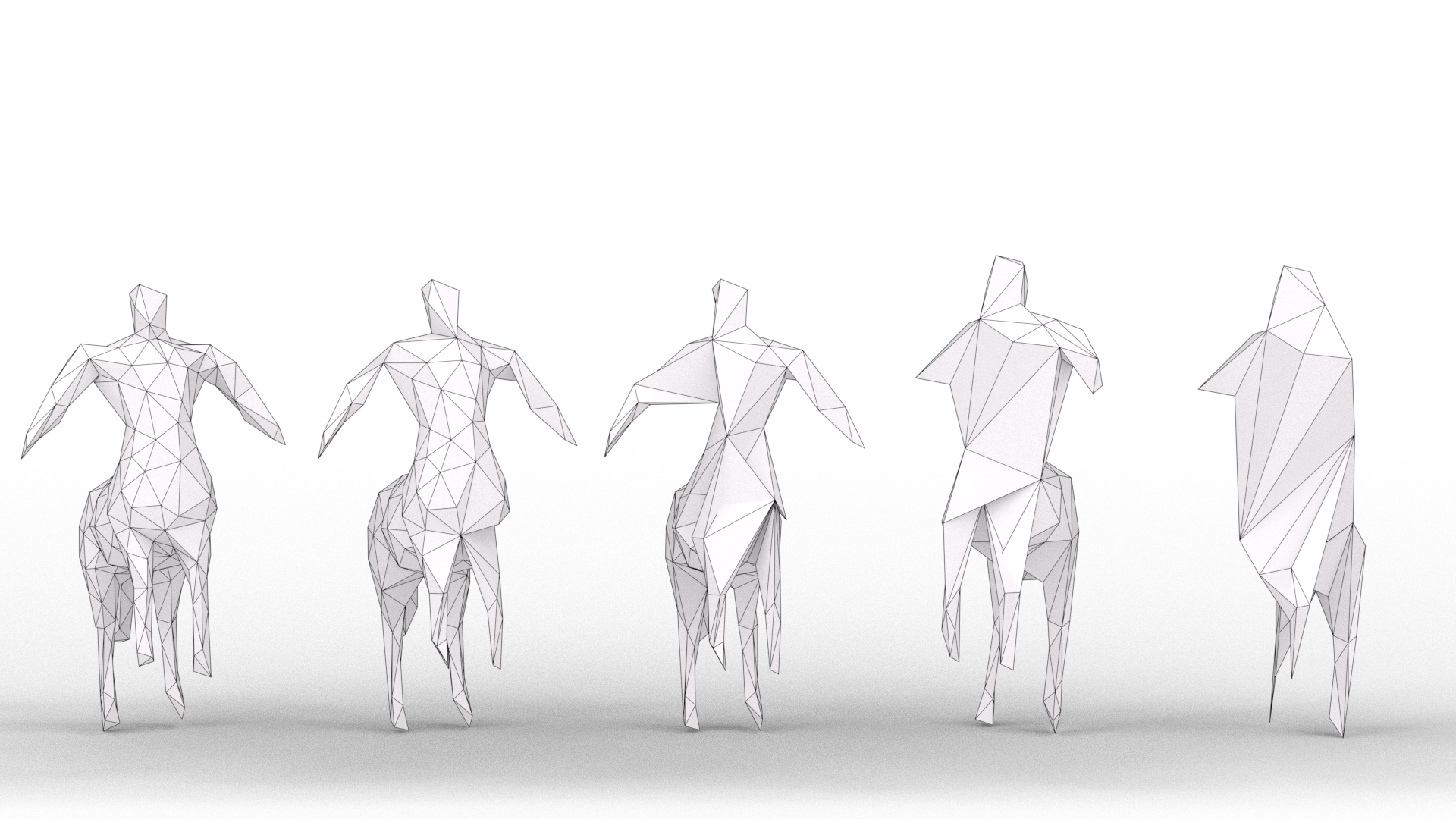} \\
\adjincludegraphics[width=\cfig cm,trim={{0\width} {0.11\height} {0.01\width} {0.36\height}},clip]{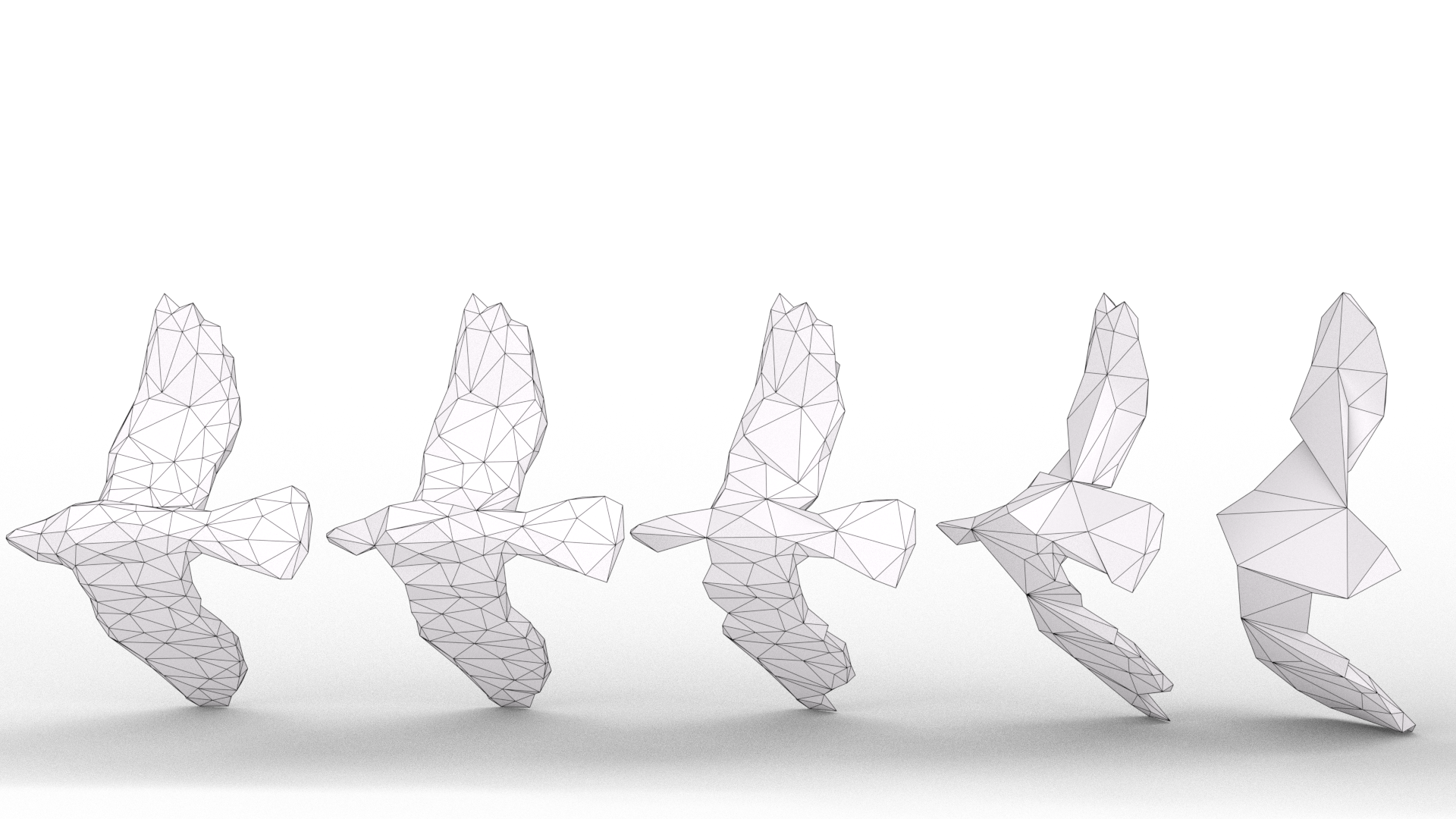} \\
\adjincludegraphics[width=\cfig cm,trim={{0\width} {0.08\height} {0\width} {0.36\height}},clip]{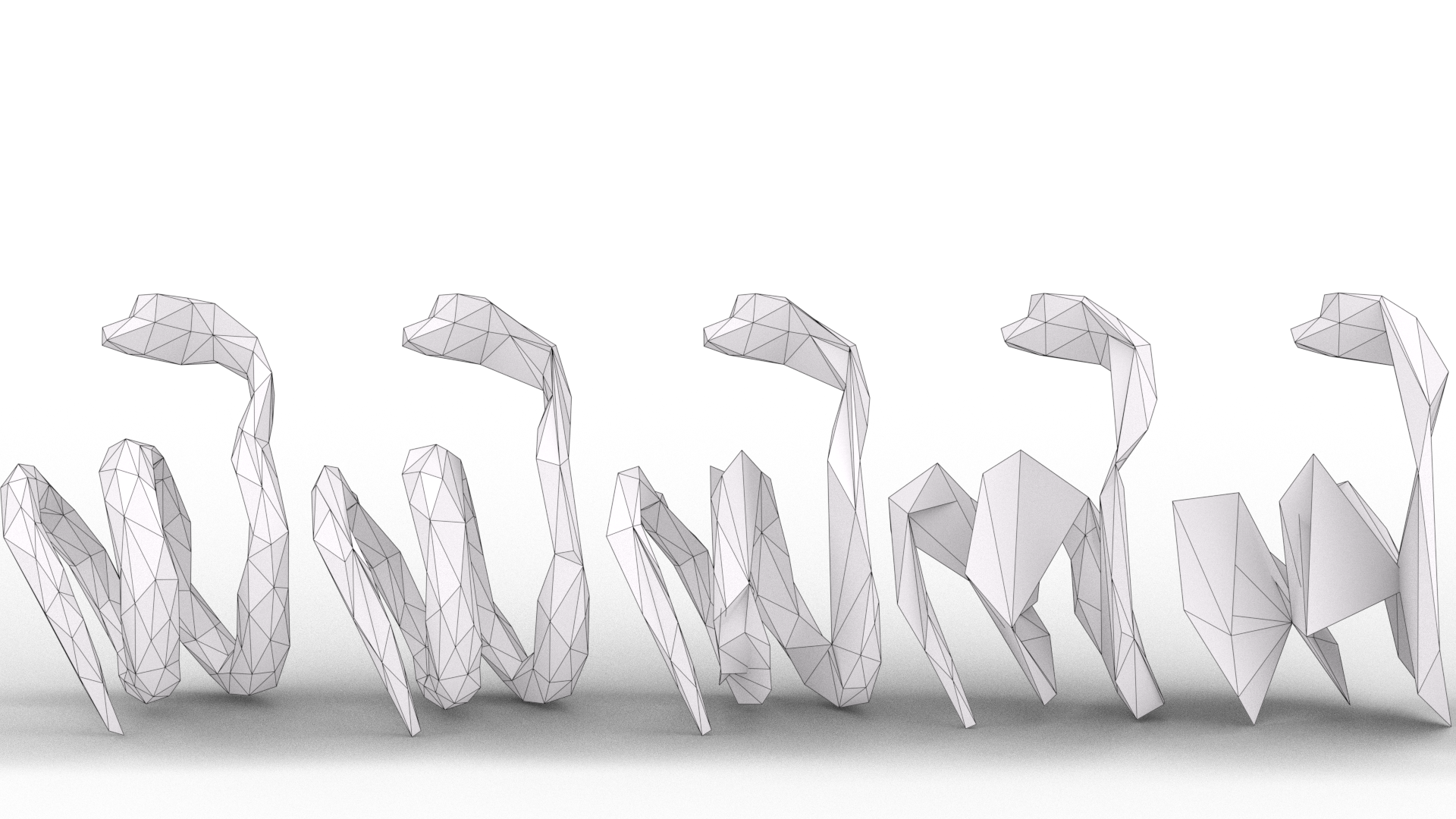} \\
\adjincludegraphics[width=\cfig cm,trim={{0\width} {0.15\height} {0\width} {0.36\height}},clip]{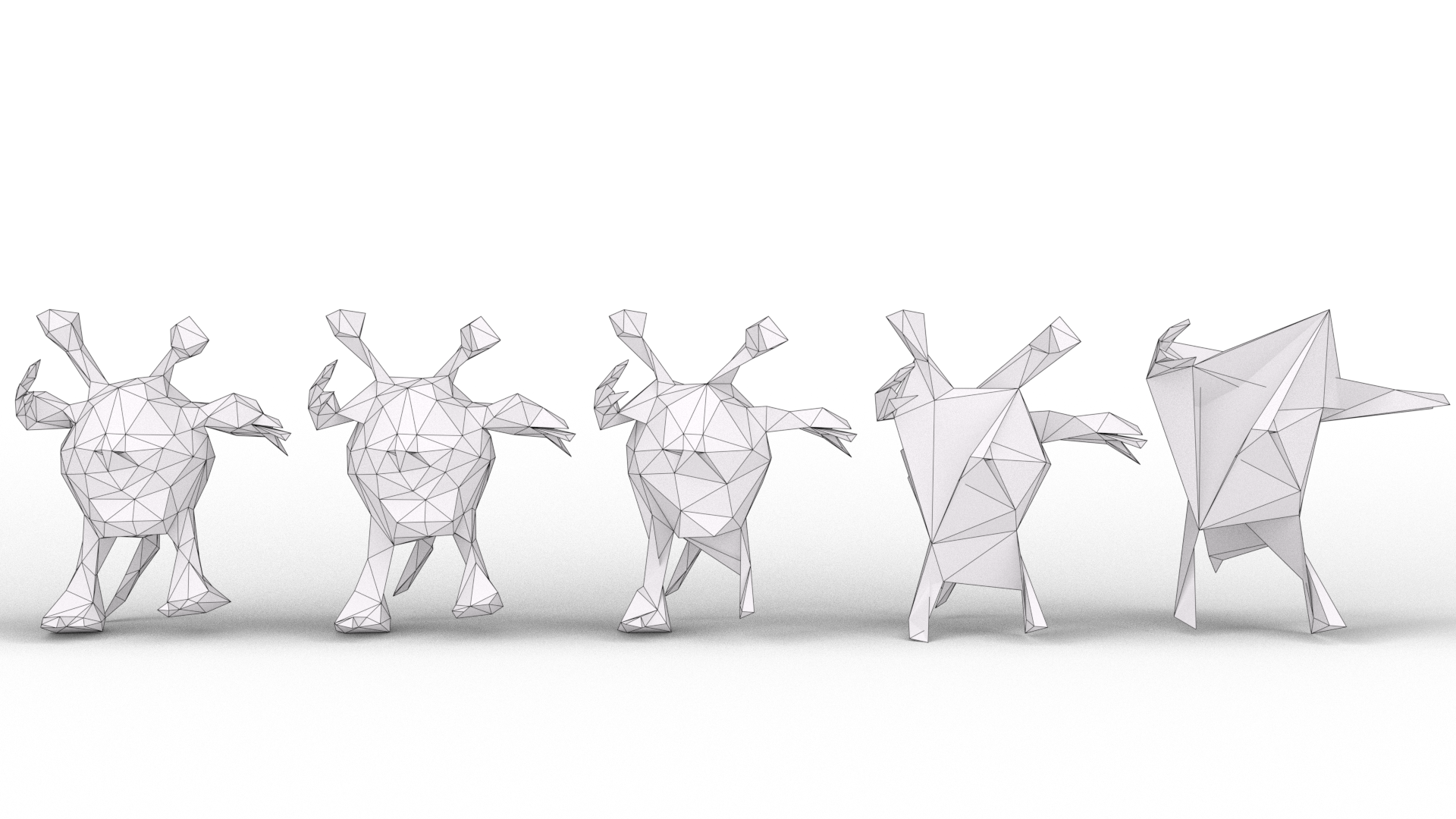} \\
\vspace{-5pt}
\begin{tabular}{C{\tfig cm} C{\tfig cm} C{\tfig cm} C{\tfig cm} C{\tfig cm}}
input & $600$ & $450$ & $300$ & $150$ \\
\end{tabular}
\end{tabular}
\end{center}
\caption{Intermediate pooled meshes on the SHREC11 classification dataset. The input meshes are all simplified to roughly 750 edges (500 faces), and are sequentially pooled to $600$, $450$, $300$ and $150$ edges.}
\label{fig:pool_shrec}
\end{figure}

\ourmethod{} is a general method for applying CNNs directly on triangular meshes, with many applications. Using the building blocks of \ourmethod{} described in Section \ref{sec:method}, we can construct different network configurations for the purpose of solving different tasks. Like conventional CNNs, these building blocks provide a \emph{plug-and-play} framework. \rg{For computational efficiency, in the pooling operation, we aggregate the features only once per pooling operation. While the edge sorting and collapse are performed sequentially, this relaxation allows performing the feature aggregation operation on a GPU, which improves the computational efficiency.}

In what follows, we demonstrate \ourmethod{} performance on classification and segmentation tasks. \rg{Details on the network architectures used are given in Appendix~\ref{sec:meshcnncfg}}.

\subsection{Data Processing}

Across all sets, we simplified each mesh to \emph{roughly} the same number of edges. Note that as mentioned earlier, \ourmethod{} does not require the same number of edges across all samples. However, similar to initial resize of images in CNNs, geometric mesh decimation helps to reduce the input resolution and with it the network capacity required for training. Since the classification task learns a global shape description, we usually use a lower resolution ($750$ edges), compared with the segmentation task ($2250$ edges).

{\bf Augmentation.} Several forms of data augmentation exist for generating more data samples for the network. Note that since our input features are similarity-invariant, applying rotation, translation and isotropic scaling (same in $x$, $y$ and $z$) does not generate new input features. However, we can generate new features by applying anisotropic scaling on the vertex locations in $x$, $y$ and $z$, $<S_x,S_y,S_z>$ (each randomly sampled from a normal distribution $\mu=1$ and $\sigma=0.1$), which \textit{will} change the input features to the network. We also shift vertices to different locations on the mesh surface.
Furthermore, we augment the tessellation of each object by performing random edge flips. Due to the flexible input resolution, we can also collapse a small random set of edges prior to training.

\subsection{Mesh Classification}
{\bf SHREC.}
\begin{table}
\begin{center}
\begin{tabular}{||c c c|| c} 
\cline{1-3}
\multicolumn{3}{c}{Classification SHREC}\\
\cline{1-3}
Method & Split $16$  & Split $10$\\ [0.2ex] 
\cline{1-3}
\textbf{MeshCNN} & $\textbf{98.6}$ & $\textbf{91.0}\%$ \\
\cline{1-3}
GWCNN & $96.6\%$ & $90.3\%$ & \rdelim\}{4}{2.2cm}[~\cite{gwcnn}] \\
GI    & $96.6\%$ & $88.6\%$ \\ 
SN    & $48.4\%$ & $52.7\%$ \\ 
SG    & $70.8\%$ & $62.6\%$ \\ 
\cline{1-3}
\end{tabular}
\end{center}
\caption{SHREC $30$ class classification (comparisons taken from~\shortcite{gwcnn}). Split $16$ and $10$ are the training splits, trained up to $200$ epochs.}
\label{tab:shrec}
\end{table}
We performed classification on $30$ classes from the SHREC dataset~\cite{shrec2011}, with $20$ examples per class. We follow the setup in~\cite{gwcnn}, where split $16$ and $10$ are the number of training examples per class and \rh{we stop training after $200$ epochs.} Since we did not have the exact splits used in~\shortcite{gwcnn}, our result is averaged over $3$ randomly generated split $16$ and $10$ sets. Table~\ref{tab:shrec} reports the results. 
\rg{For comparison,} we take the evaluations directly from~\shortcite{gwcnn}, who compare against: SG~\cite{bronstein2011shape} (bag-of-features representation), SN~\cite{Wu15Shapenet} (volumetric CNN), GI~\cite{sinha2016deep} (CNN on fixed geometry images) and finally GWCNN~\shortcite{gwcnn} (learned geometry images). \rg{The advantage of our method is apparent. }
We visualize \rg{some examples of} mesh pooling simplifications  \rg{of this dataset} in Figure~\ref{fig:pool_shrec}. We observe that mesh pooling behaves in a consistent \emph{semantic} manner (see Figure~\ref{fig:task_pool_shrec}).

\begin{figure}
\adjincludegraphics[width=8cm,trim={{0.04\width} {0.0\height} {0.07\width} {0\height}},clip]{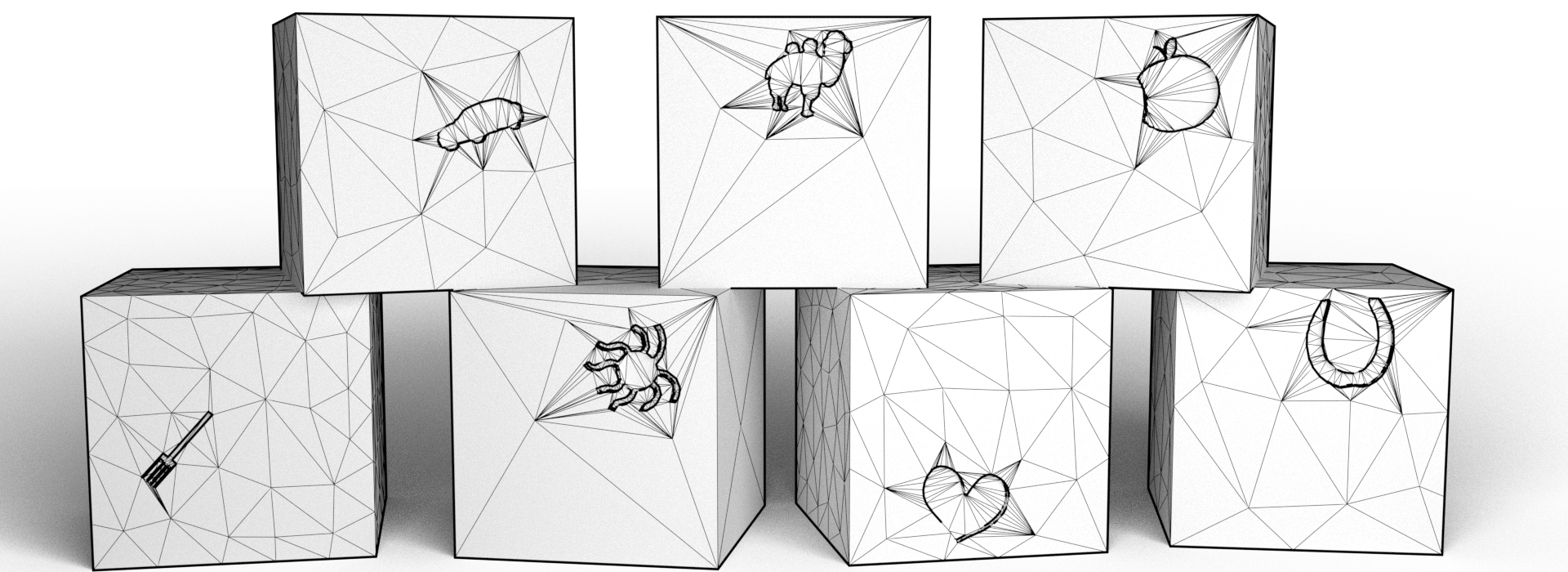} 
%
\caption{Engraved cubes classification dataset. We generate $23$ different classes (\emph{e.g.,} car, heart, apple, etc.) by extruding stickers from MPEG-7~\cite{latecki2000shape}, and placing them on a random face in a random location.}
\label{fig:cube_figures}
\end{figure}
{\bf Cube engraving.} 
To illustrate the distinctive power of MeshCNN, we modeled a set of cubes with shallow icon engravings (see Figure~\ref{fig:cube_figures}). 
We use $23$ classes from the MPEG-7 binary shape~\cite{latecki2000shape} dataset, with roughly $20$ icons per class. We set aside three icons per class for the test set, and use the remainder for training. For each icon, we randomly sample $10$ different locations (position, rotation and cube face) to inset the icon. Each cube has approximately $500$ faces, which means that detailed shapes have fewer triangles in the flat areas, while less detailed shapes have more triangles in flat areas. This set contains a total of $4600$ shapes with $3910$ / $690$ in the train / test split.
We plan to release this dataset as well as the data synthesis code after publication.

\begin{figure}
\newcommand{\cfig}{8.3}
\begin{tabular}{c}
\adjincludegraphics[width=\cfig cm,trim={{0.0\width} {0.1\height} {0\width} {0.15\height}},clip]{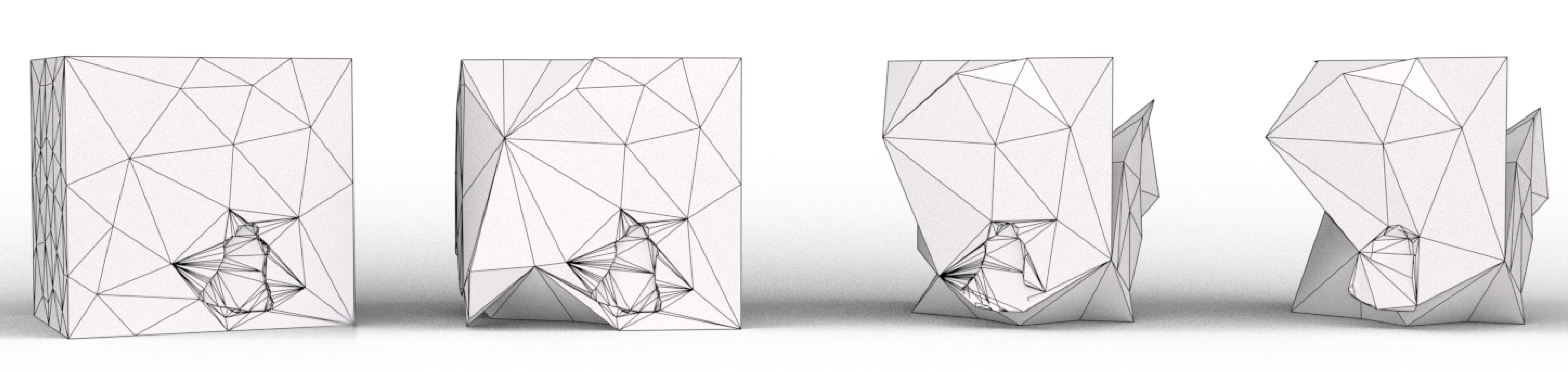} \\
\adjincludegraphics[width=\cfig cm,trim={{0.0\width} {0.1\height} {0\width} {0.15\height}},clip]{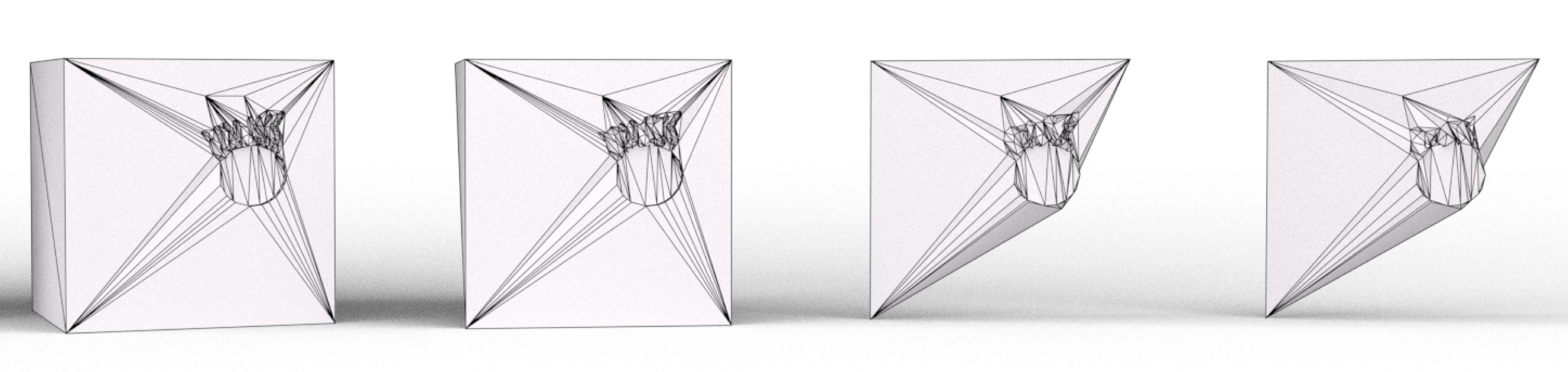} \\
\adjincludegraphics[width=\cfig cm,trim={{0.0\width} {0.1\height} {0\width} {0.15\height}},clip]{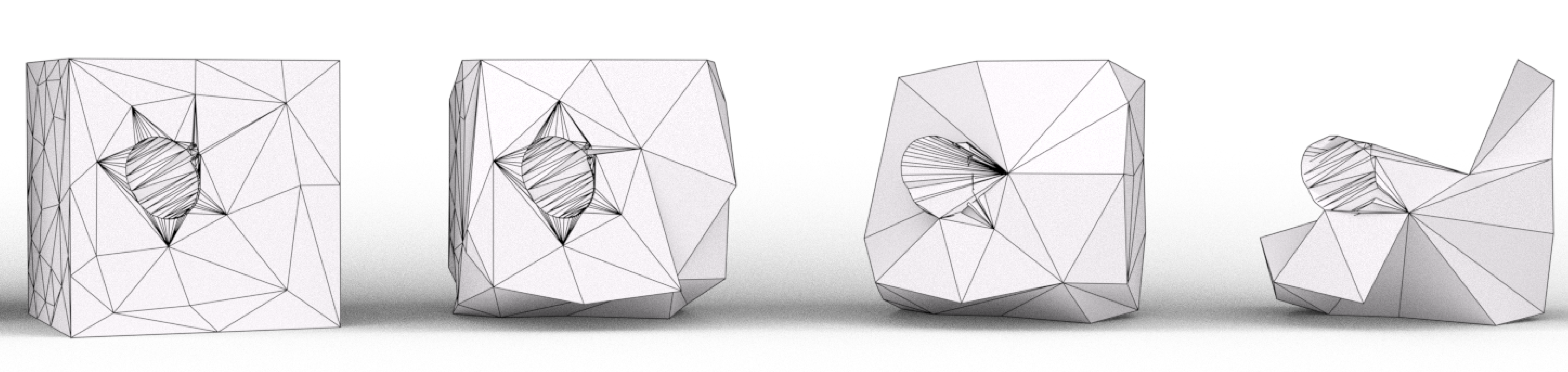} \\
\adjincludegraphics[width=\cfig cm,trim={{0.0\width} {0.1\height} {0\width} {0.15\height}},clip]{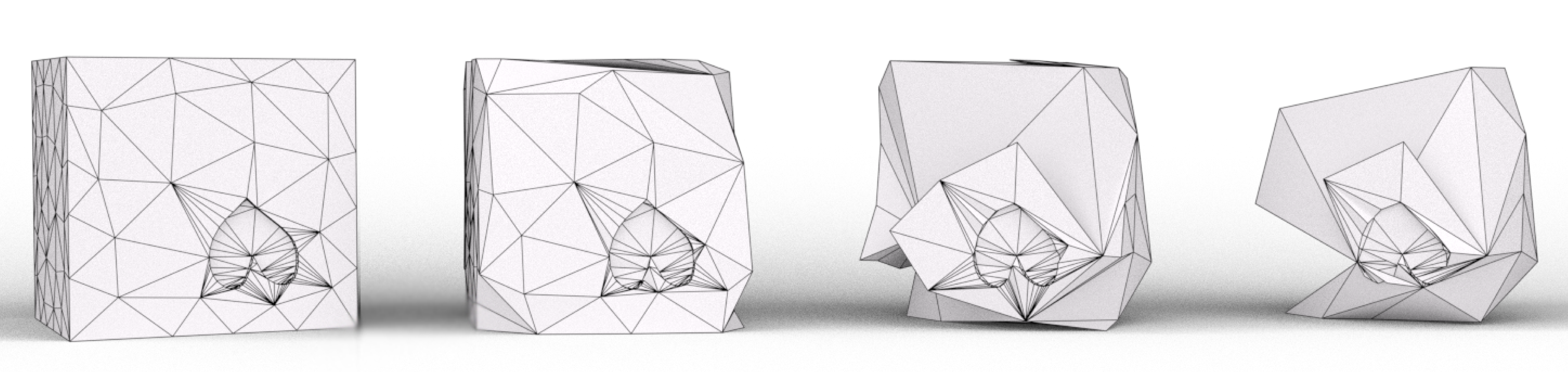} \\
\end{tabular}
\caption{\ourmethod{} trained to predict the class of icon engraving. Observe how the network learns to preserve important edges and remove redundant edges with regards to the classification task.}
\label{fig:pool_cubes}
\end{figure}

We train \ourmethod{} to classify the cubes. We show the quantitative results in Table~\ref{tab:syn_cube}. To visualize the effect of mesh pooling on the classification task, we extracted the intermediate results following each mesh pooling operation (shown in Figure~\ref{fig:pool_cubes}). Observe how \ourmethod{}  learned to reduce the edges irrelevant to the classification task (flat cube surfaces) while preserving the edges within and surrounding the icon engraving.

We also trained point-based approaches on this set and show results in Table~\ref{tab:syn_cube}. 
\rhc{danny please check this... its in comment for now...}
\rhc{Note that we tried several different sampling settings. Namely, there are only $252$ vertices on the $500$ face cubes, which are not ideally placed on the mesh contour engraving. The best setting we found for PointNet++ was with uniform sampling and complementing with the mesh face normals.}
While this example may be considered contrived, it is meant to highlight that \ourmethod{} excels on 3D shapes that contain 
\dc{a large variance in geometric resolution.}  
\begin{table}
\begin{center}
\begin{tabular}{||c c c||} 
\hline
\multicolumn{3}{c}{Cube Engraving Classification}\\
\hline
method & input res & test acc \\ [0.2ex] 
\hline\hline\hline\hline
MeshCNN & $750$ & $\textbf{92.16}\%$ \\ 
\hline\hline\hline\hline
PointNet++ & $4096$ & $64.26\%$ \\ 
\hline
\end{tabular}
\end{center}
\caption{Results on engraved cubes (shown in Figure~\ref{fig:cube_figures}). \rhc{The fine grained details are difficult to perceive using only Cartesian coordinates. Incorporating the surface normal information.}}
\label{tab:syn_cube}
\end{table}



\subsection{Mesh Segmentation}
Another application of \ourmethod{} is consistent shape segmentation, which is an important building block for many applications in shape analysis and synthesis.
We used supervised learning to train \ourmethod{} to predict, for every edge, the probability of belonging to a particular segment on the COSEG~\cite{wang2012active} and Human Body Segmentation~\cite{maron2017convolutional} datasets. \rhc{Training configurations in Appendix~\ref{sec:meshcnncfg}.}
Since both datasets provide ground truth segmentation per face, we generated edge-level semantic labeling on the simplified meshes based on the labels from the original resolution.

The most straightforward \ourmethod{} semantic segmentation configuration would be to use a sequence of mesh convolution layers (along with normalization and non-linear activation units). However, incorporating mesh pooling enables \ourmethod{} to learn a segmentation-driven edge collapse.
Recall that mesh pooling reduces the input mesh resolution, which is no longer consistent with the ground truth edge-level labels. 
To this end, we use the mesh unpooling layer to upsample the resolution back to the original input size.

{\bf COSEG.}
We evaluate the performance of \ourmethod{} on the task of segmentation on the COSEG dataset, which contains three large sets: \emph{aliens}, \emph{vases} and \emph{chairs} containing $200$, $300$ and $400$ models in each respectively. We split each shape category into $85 \% / 15 \%$ train/test splits. \rg{We compare to PointNet, PointNet++ and PointCNN and report the best accuracy for all methods in Table \ref{tab:cosegtbl}.
Our technique achieves better results than all the other methods on this dataset.}

\begin{figure}
\adjincludegraphics[width=8.3 cm,trim={{0.0\width} {0.0\height} {0.0\width} {0.0\height}},clip]{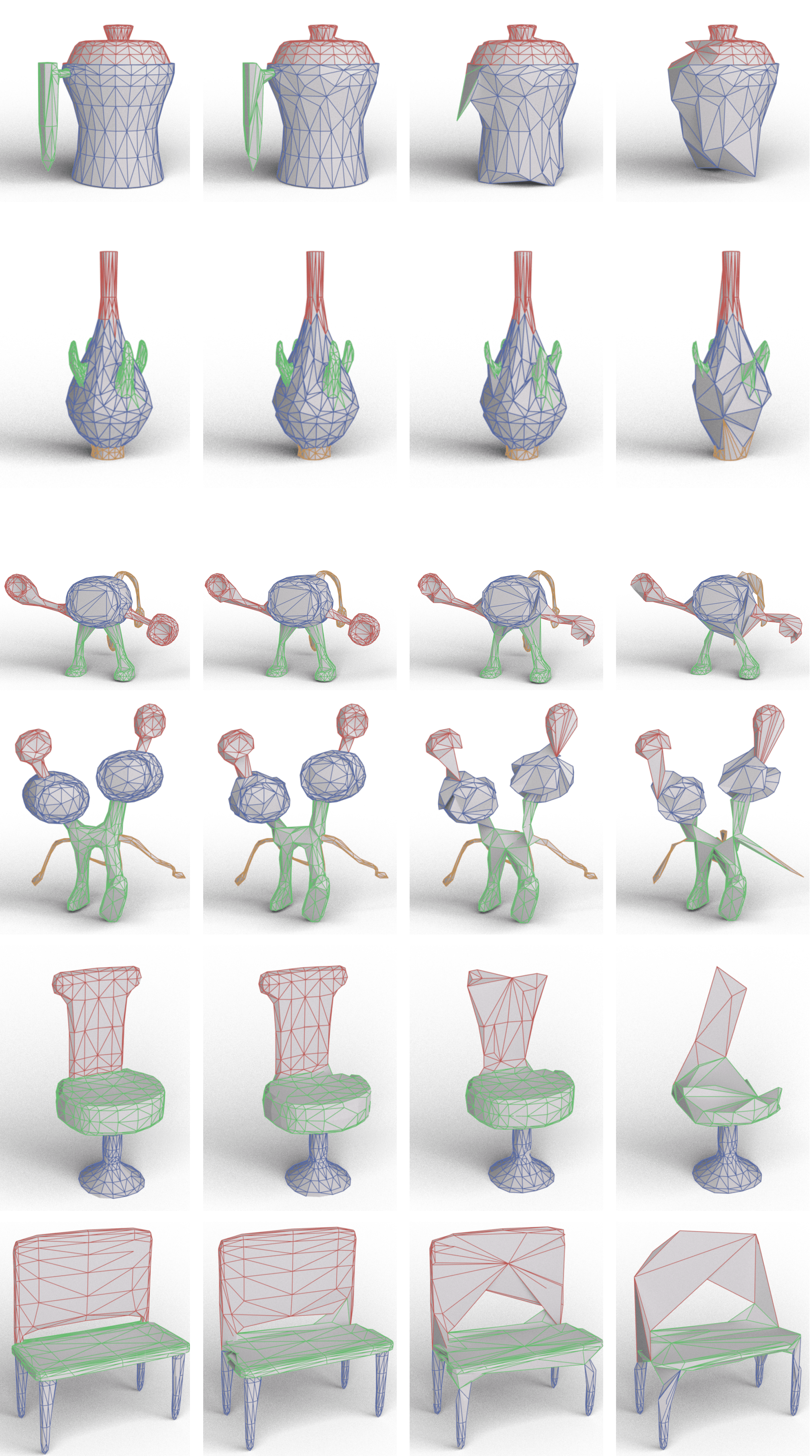}
\caption{Semantic segmentation results on (held out) test shapes. The segmentation prediction per edge is shown on the left, followed by the intermediate simplified meshes after each pooling layer. For visualization purposes, the edges in the intermediate meshes are colored with the final segmentation predictions. \ourmethod{} uses mesh pooling to learn to collapse edges from the same segments, which are unrolled back to the original input mesh resolution via the mesh unpooling layer (for example, see the top row, where the entire vase handle has collapsed to a single edge).}
\label{fig:coseg}
\end{figure}

\rg{We believe that this is due to the fact that our network is tailored to the mesh structure, which gives it an advantage over the other strategies. To further show this, we report also the results for the case of random pooling (collapsed edges are picked randomly)  and exhibit that this change reduces the performance of the network. }
In addition, the final segmentation predictions from \ourmethod{} semantic segmentation network with pooling and unpooling layers on a held out test set is shown in Figure~\ref{fig:coseg}. This also manifests how the pooling performed is adapted to the  target problem. 
\begin{table}
\begin{center}
\begin{tabular}{||c c c c||} 
\hline
\multicolumn{4}{c}{COSEG Segmentation }\\
\hline
Method & Vases & Chairs & Telealiens \\ [0.2ex] 
\hline\hline\hline\hline\hline
MeshCNN (\textit{UNet}) & $\textbf{97.27}\%$ & $\textbf{99.63}\%$ & $\textbf{97.56}\%$ \\
MeshCNN (\textit{rand. pool}) & $96.64\%$ & $99.23\%$ & $97.01\%$ \\ 
\hline\hline\hline\hline
PointNet & $91.5\%$ & $70.2\%$ &  $54.4\%$\\ 
PointNet++ & $94.7\%$ & $98.9\%$ &  $79.1 \%$\\ 
PointCNN & $ 96.37 \%$ & $ 99.31 \%$ &  $ 97.40 \%$\\ 
\hline
\end{tabular}
\end{center}
\caption{\rh{MeshCNN evaluations on COSEG segmentation. \rhc{TODO}}}
\label{tab:cosegtbl}
\end{table}

\begin{figure}
\newcommand{\cfig}{2.5}
\begin{tabular}{cccc}
\adjincludegraphics[height=\cfig cm,trim={{0.46\width} {0.15\height} {0.35\width} {0.05\height}},clip]{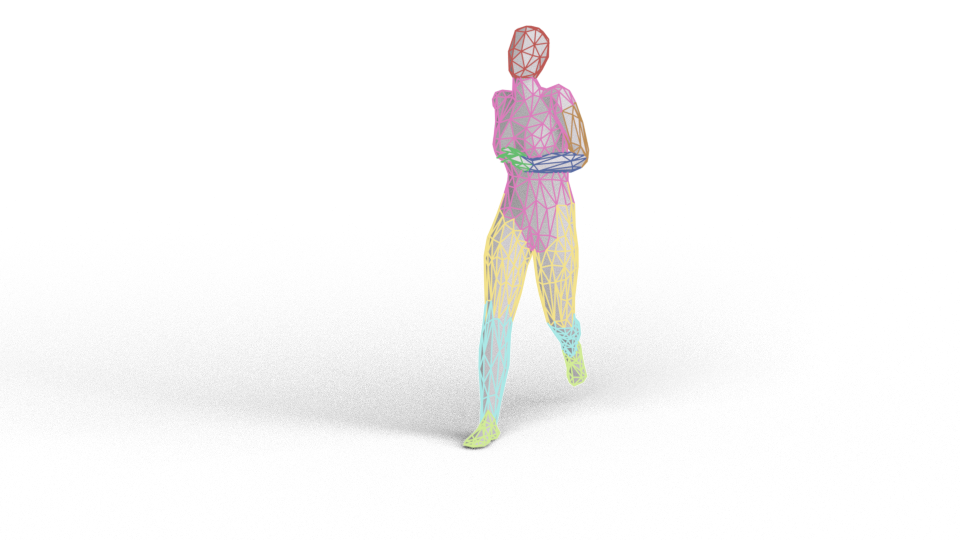}  &
\adjincludegraphics[height=\cfig cm,trim={{0.4\width} {0.15\height} {0.28\width} {0.05\height}},clip]{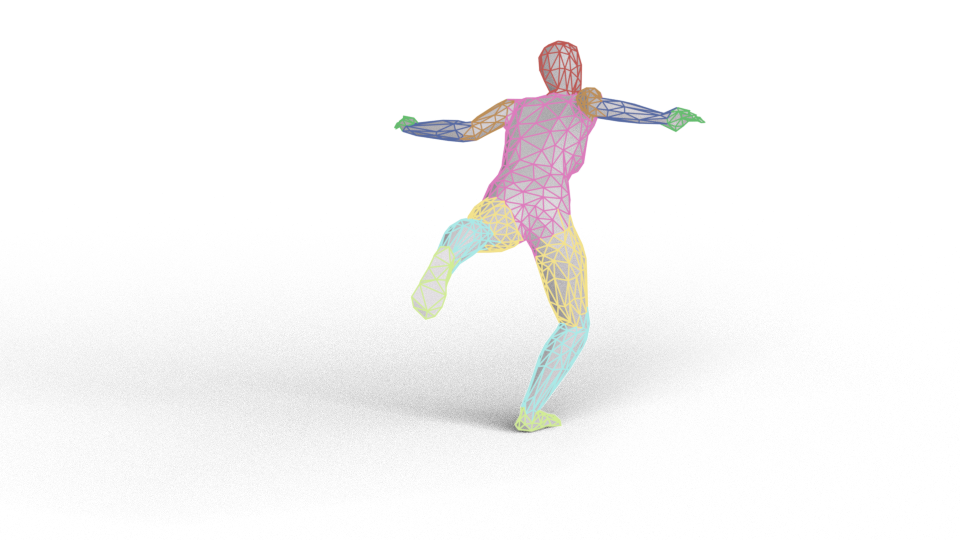} &
\adjincludegraphics[height=\cfig cm,trim={{0.42\width} {0.15\height} {0.35\width} {0.05\height}},clip]{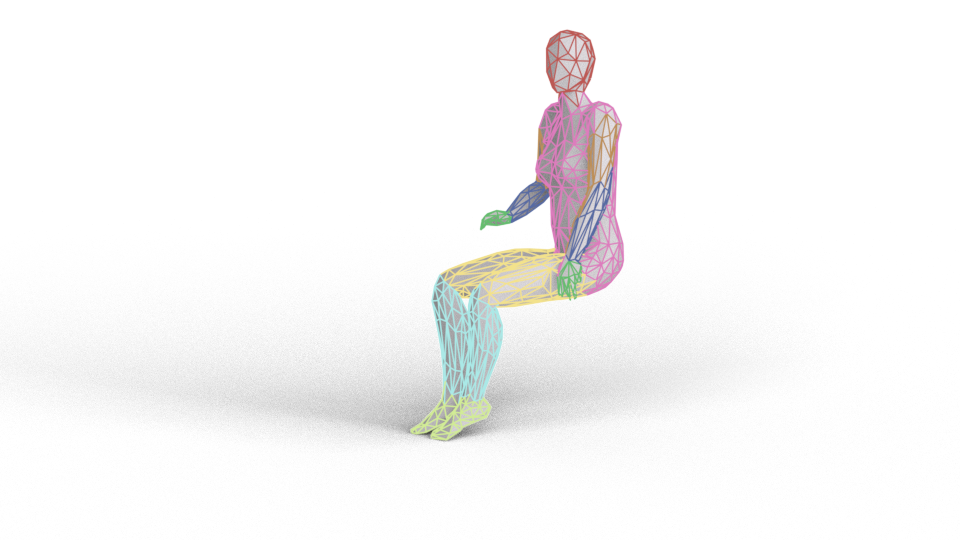} &
\adjincludegraphics[height=\cfig cm,trim={{0.4\width} {0.15\height} {0.3\width} {0.05\height}},clip]{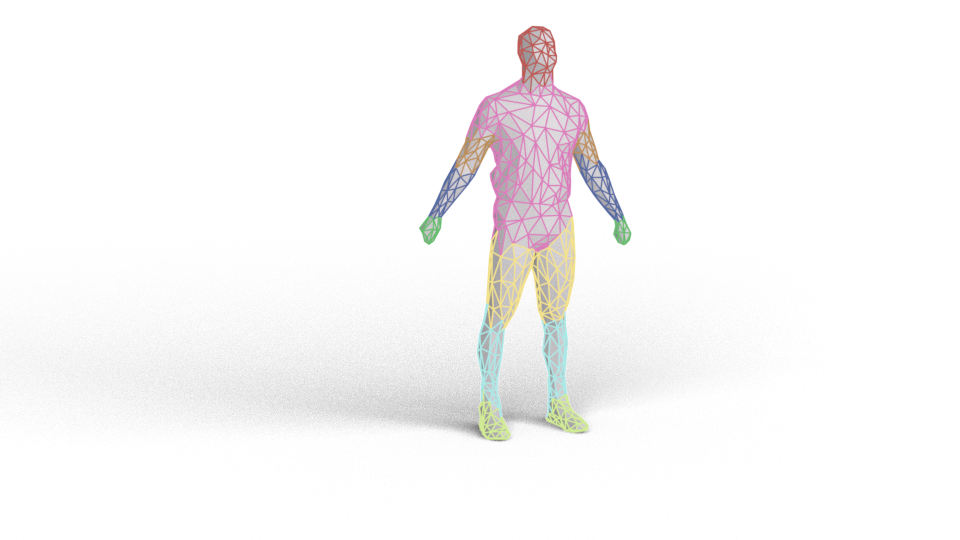} \\
\end{tabular}
\caption{Human shape segmentation results on the dataset of~\cite{maron2017convolutional}.}
\label{fig:human_seg}
\end{figure}

\begin{figure}
\newcommand{\cfig}{8.3}
\begin{center}
\adjincludegraphics[width=\cfig cm,trim={{0.03\width} {0.04\height} {0.04\width} {0.04\height}},clip]{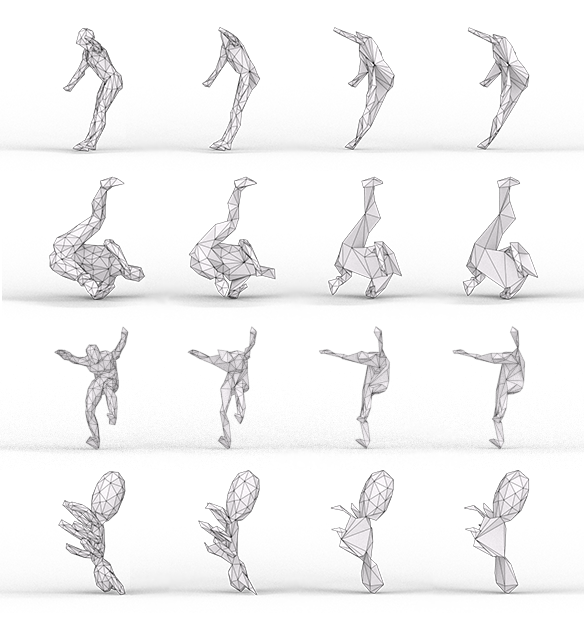}
\end{center}
\caption{\rh{Adaptive pooing has the potential to expose semantics within a class. We observe a consistent semantic pooling within the same class on the task of shape classification (SHREC). For example, the heads of people are analogously pooled, while similar shape attributes are pooled differently for another class (bottom).}}
\label{fig:task_pool_shrec}
\end{figure}

{\bf Human Body Segmentation.}
We evaluated our method on the human body segmentation dataset proposed by~\cite{maron2017convolutional}. The dataset consists of $370$ training models from SCAPE~\cite{anguelov2005scape}, FAUST~\cite{bogo2014faust}, MIT~\cite{vlasic2008articulated} and Adobe Fuse~\cite{adobe}, and the test set is $18$ models from SHREC07~\cite{giorgi2007shape} (humans) dataset. The models are manually segmented into eight labels according to the labels in~\cite{kalogerakis2010learning}. Recently,~\cite{Poulenard} reported results on this dataset for their method with comparisons to GCNN~\cite{Masci15Geodesic}, PointNet++~\cite{Qi17PointNetpp}, Dynamic Graph CNN~\cite{wang2018dynamic}, and Toric Cover~\cite{maron2017convolutional}. We take the reported results directly from ~\cite{Poulenard} and list them in Table~\ref{tab:humanseg}.
\dc{We added to the table the recent results by Haim et al. \shortcite{Haim2018Surface}, reporting state-of-the-art results on this set. Also in this case, \ourmethod{} has an advantage over the other methods (part are graph/manifold based and some are point based), which we believe results from the adaptivity of \ourmethod{} both to the mesh structure and the task at hand. Figure~\ref{fig:human_seg} presents some qualitative results of \ourmethod{}.}

\begin{table}
\begin{center}
\begin{tabular}{||c c c|| c} 
\cline{1-3}
\multicolumn{3}{c}{Human Body Segmentation}\\
\cline{1-3}
Method & \# Features & Accuracy \\ [0.2ex] 
\cline{1-3}
MeshCNN & $5$ & $\textbf{92.30} \%$ \\ 
\cline{1-3}
SNGC         &  $3$ & $91.02\%$ & \\ 
Toric Cover & $26$ & $88.00\%$ & \rdelim\}{5}{1cm}[~\shortcite{Poulenard}] \\
PointNet++  &  $3$ & $90.77\%$ \\ 
DynGraphCNN &  $3$ & $89.72\%$ \\ 
GCNN       &  $64$ & $86.40\%$ \\ 
MDGCNN     &  $64$ & $89.47\%$ \\ 
\cline{1-3}
\end{tabular}
\end{center}
\caption{Human body segmentation results (cited comparisons taken from~\cite{Poulenard}). \rh{Latest comparison taken from~\cite{Haim2018Surface}}. }
\label{tab:humanseg}
\end{table}

\subsection{Additional Evaluations}
{\bf Computation Time.}
\rh{Our non-optimized PyTorch~\cite{paszke2017automatic} implementation takes an average of $0.21 / 0.13$ seconds per example when training on segmentation / classification with $2250 / 750$ edges using a GTX $1080$ TI graphics card.}

\begin{figure}
\newcommand{\cfig}{1.7}
\begin{tabular}{cccc}
\adjincludegraphics[width=\cfig cm,trim={{0.36\width} {0.12\height} {0.28\width} {0.12\height}},clip]{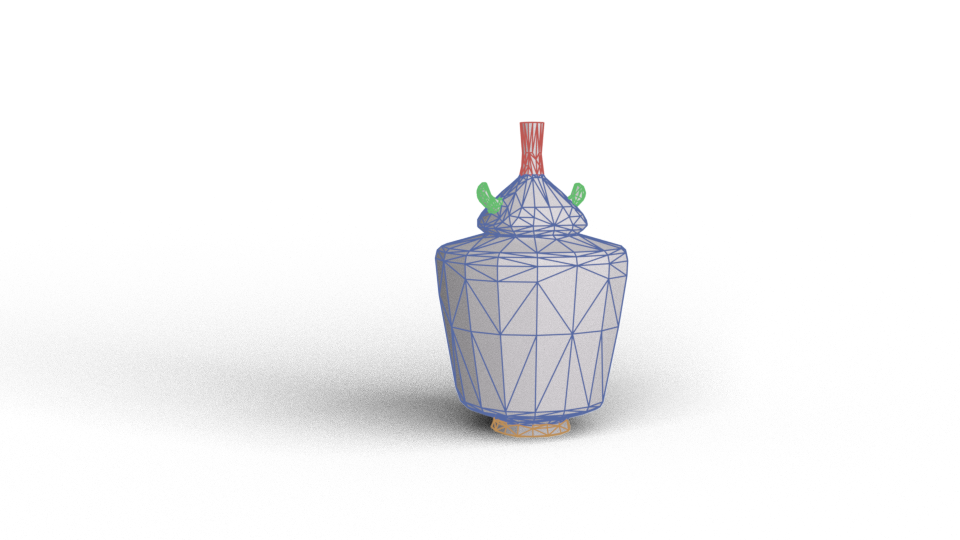} &
\adjincludegraphics[width=\cfig cm,trim={{0.36\width} {0.12\height} {0.28\width} {0.12\height}},clip]{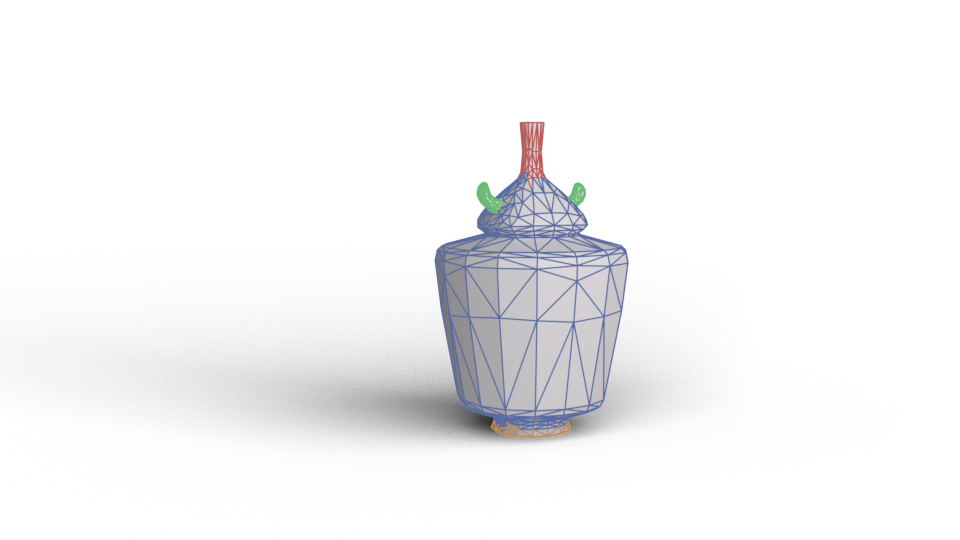} &
\adjincludegraphics[width=\cfig cm,trim={{0.3\width} {0.12\height} {0.2\width} {0.12\height}},clip]{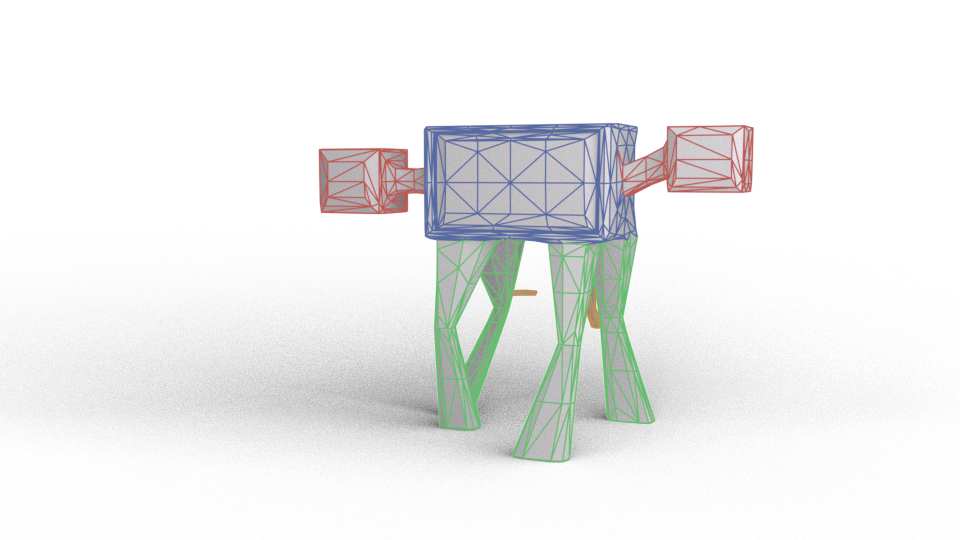} &
\adjincludegraphics[width=\cfig cm,trim={{0.3\width} {0.12\height} {0.2\width} {0.12\height}},clip]{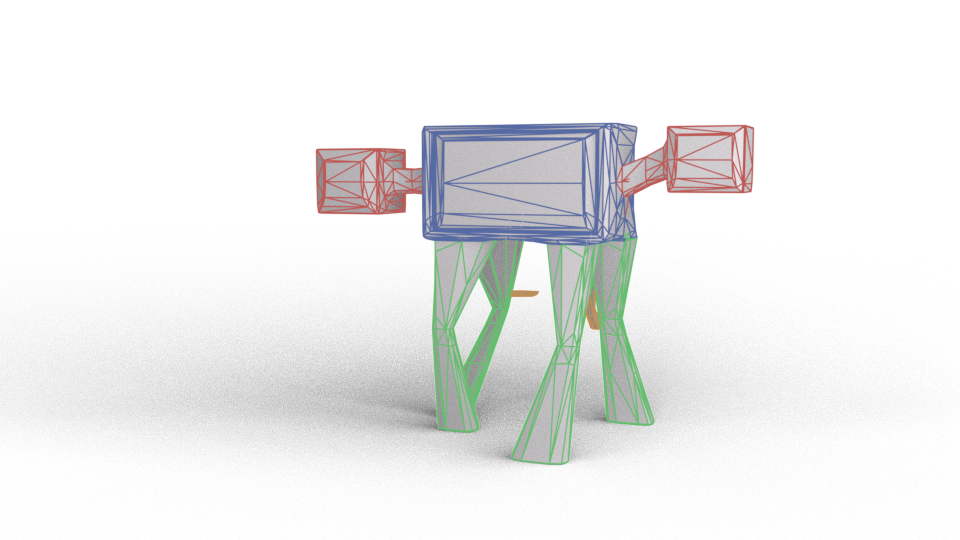} \\
\adjincludegraphics[width=1.5 cm,trim={{0.36\width} {0.08\height} {0.28\width} {0.08\height}},clip]{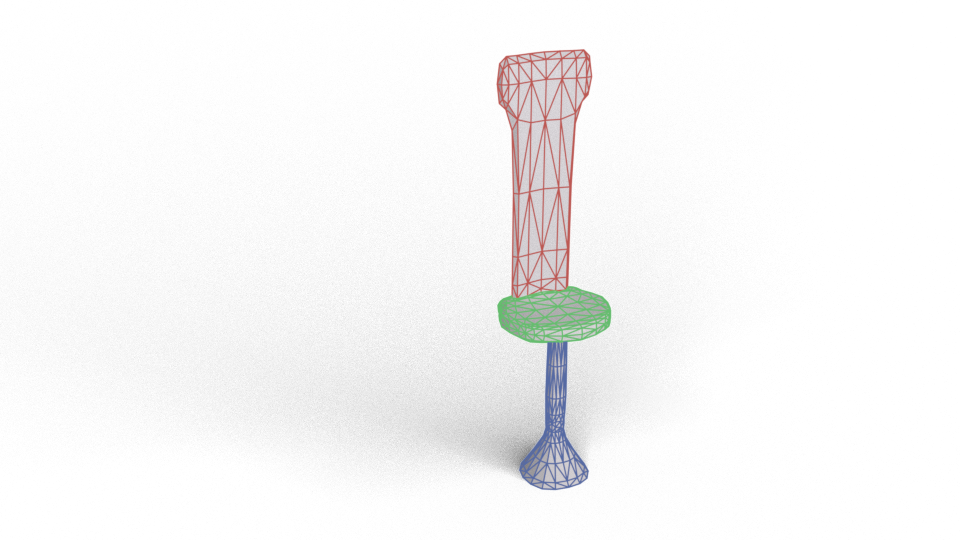} &
\adjincludegraphics[width=1.5 cm,trim={{0.36\width} {0.08\height} {0.28\width} {0.08\height}},clip]{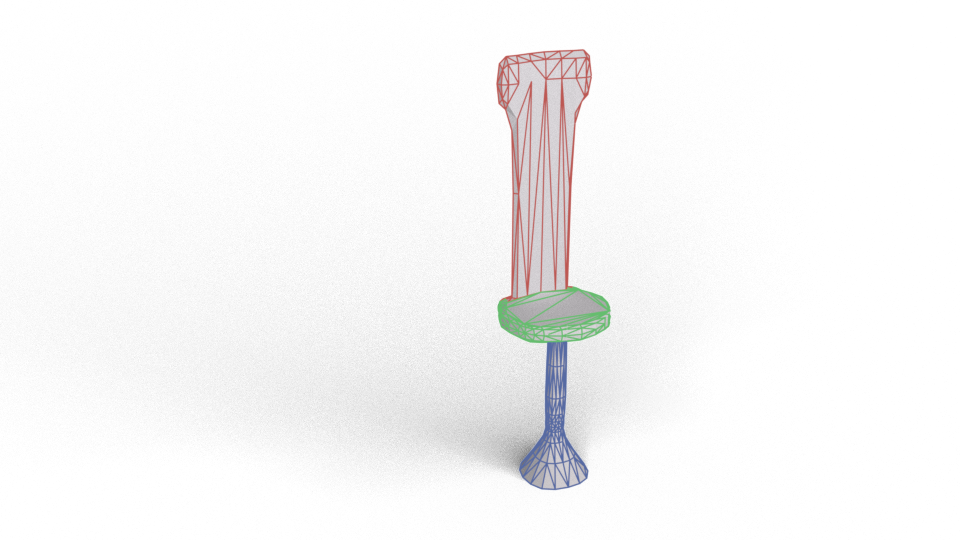} &
\adjincludegraphics[width=\cfig cm,trim={{0.36\width} {0.12\height} {0.28\width} {0.42\height}},clip]{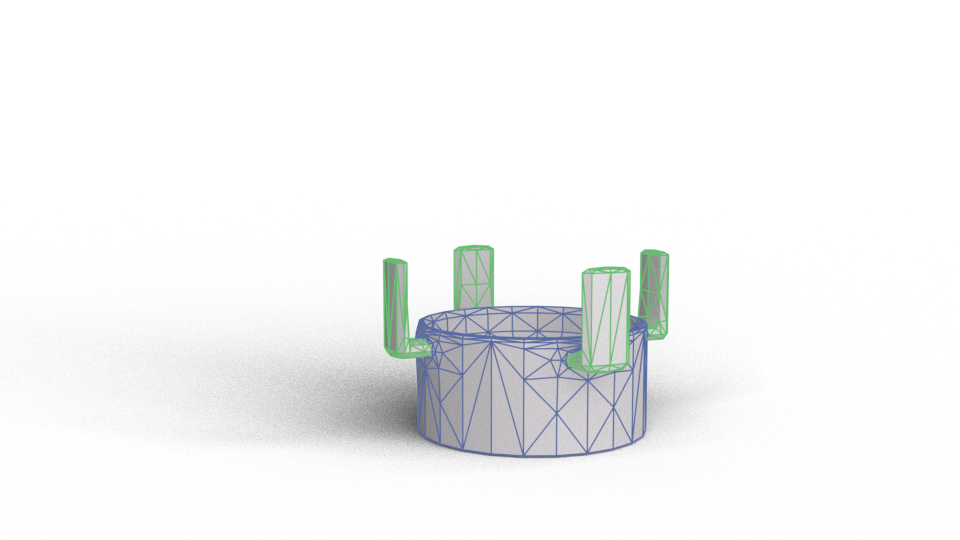} &
\adjincludegraphics[width=\cfig cm,trim={{0.36\width} {0.12\height} {0.28\width} {0.42\height}},clip]{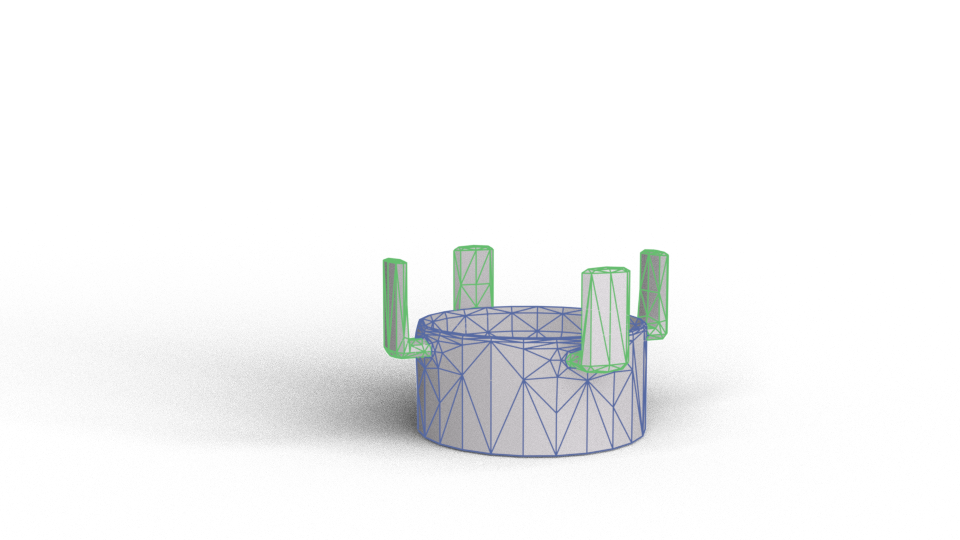} \\
\end{tabular}
\caption{Robustness to different triangulations (semantic segmentation). Left: test shapes, right: alternate triangulation.}
\label{fig:tri_robust}
\end{figure}

{\bf Tessellation Robustness.}
We examine the robustness of our method to differences in triangulations through several qualitative and quantitative experiments using the COSEG segmentation dataset.
\rhc{Todo: write about quantitative tessellation \ input features}
\nf{To this end, we generate two modified versions of the dataset. The first is obtained by applying a remeshing procedure (using Blender and MeshLab), and the second by randomly perturbing $30\%$ of the vertex positions, realized as a shift towards a random vertex in its \onering{}. The minor differences in performance (see Table~\ref{tab:robust}) imply resiliency to tessellation variation.}
See Figure~\ref{fig:tri_robust} for qualitative results.

\begin{table}
\begin{center}
\begin{tabular}{||c c c c||} 
\hline
\multicolumn{4}{c}{Robustness}\\
\hline
Set & Vases & Chairs & Aliens \\ [0.2ex] 
\hline
Reference & $97.27$ & $99.63\%$ & $97.56 \%$\\ 
\hline\hline\hline\hline
Remeshing & $96.33 \%$ & $88.98 \%$  & $97.77 \%$ \\ 
Perturbations & $97.12 \%$ & $99.60 \%$ & $97.34\%$ \\ 
\hline
\end{tabular}
\end{center}
\caption{\rh{Quantitative results on two robustness sets.} \rhc{todo..}}
\label{tab:robust}
\end{table}


\nf{
\begin{figure}
\setlength{\tabcolsep}{1pt}
\begin{tabular}{ccc}
`
\includegraphics[width=5cm]{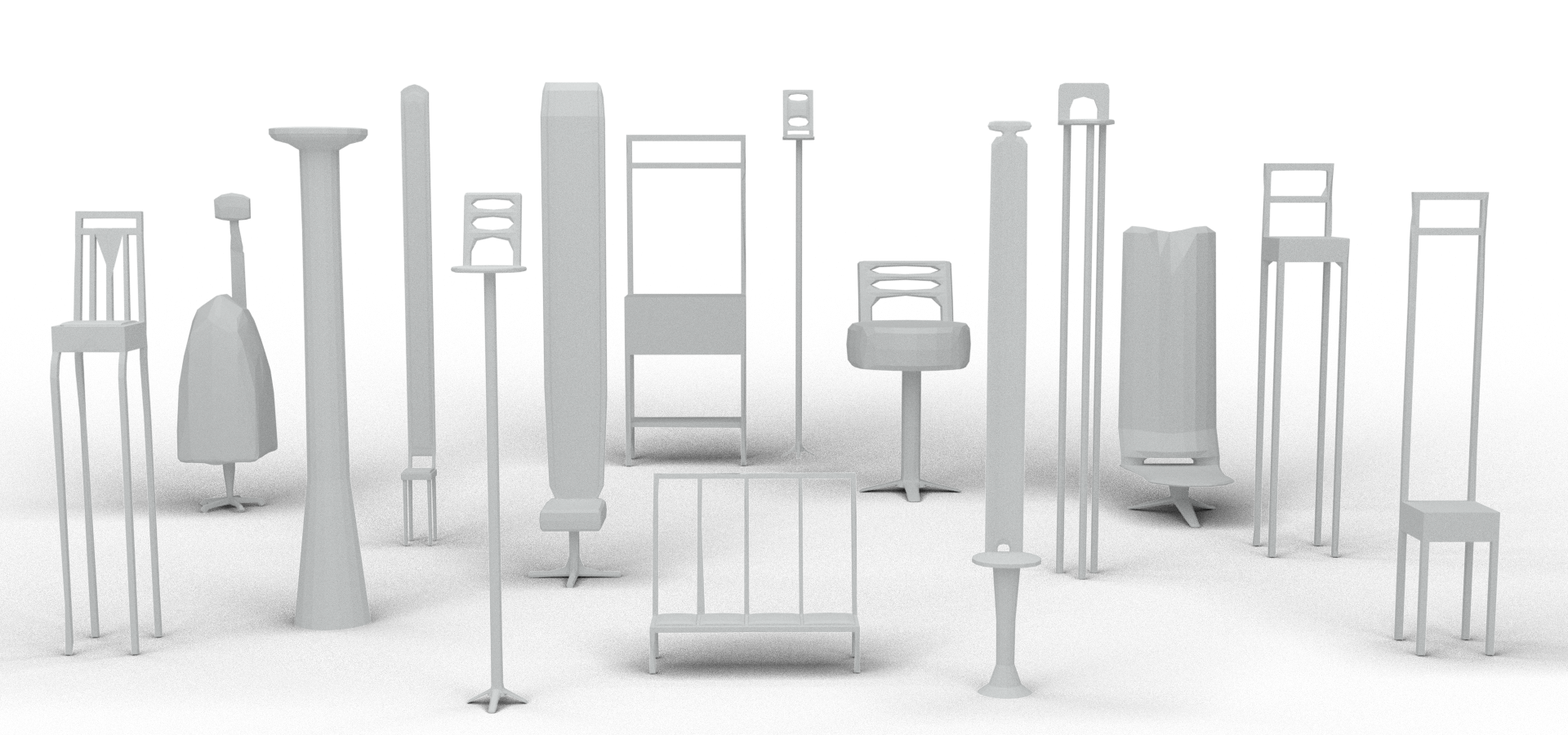} &
\adjincludegraphics[width=1.2 cm,trim={{0.45\width} {0.0\height} {0.3\width} {0.0\height}},clip]{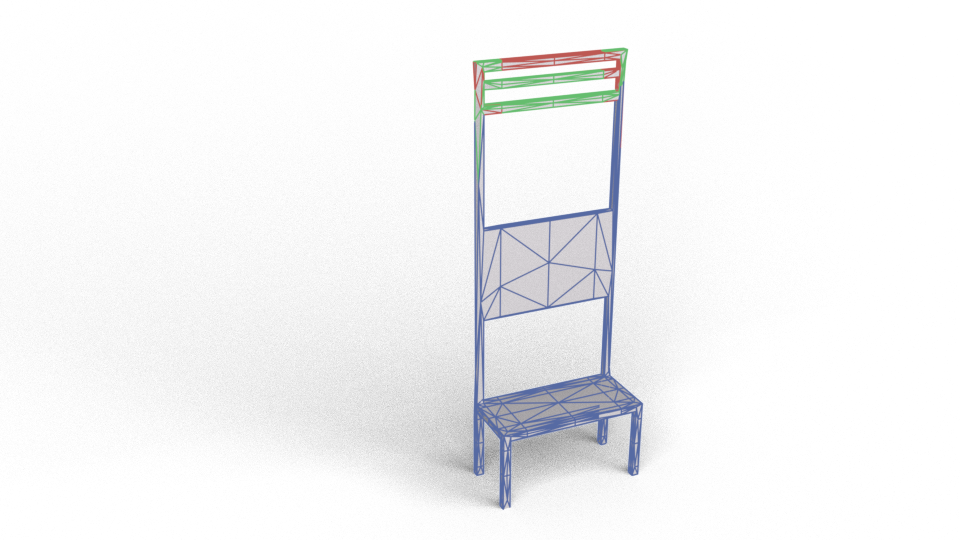} &
\adjincludegraphics[width=1.2 cm,trim={{0.45\width} {0.0\height} {0.3\width} {0.0\height}},clip]{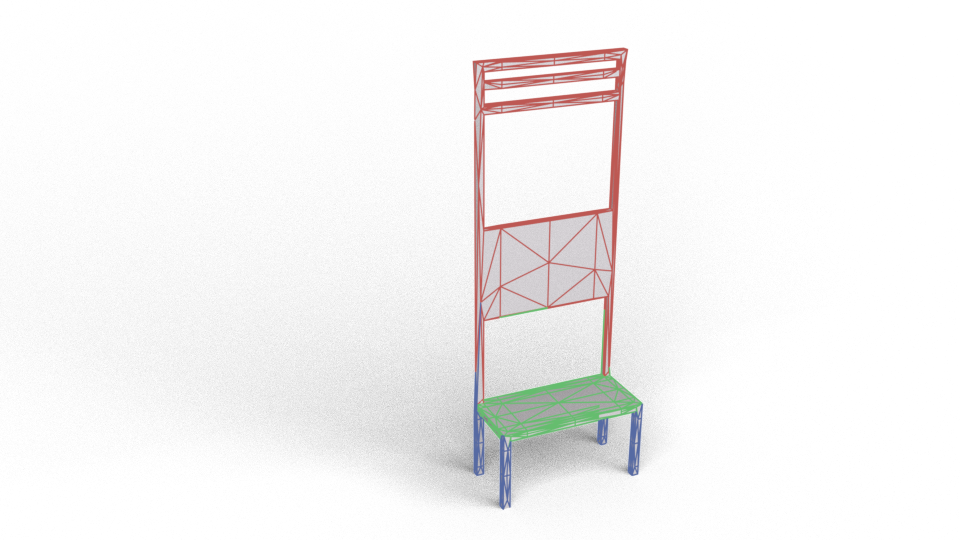} \\
Stretched Set & 3D points  & geometric  \\
\end{tabular}
\caption{\rh{Robustness of differential features. We evaluate generalization on stretched set (left), using \ourmethod{} with geometric features compared to standard $3$D points (edge midpoints). While both achieve high accuracy on the standard test set, \ourmethod{} geometric features (right) generalizes well to the stretched set compared  to the 3D points case (middle).}}
\label{fig:strech_chair_set}
\end{figure}

{\bf Invariant Features.}
A noteworthy advantage of working with relative features is that \ourmethod{} is guaranteed to be invariant to rotations, translations and uniform scaling. Inherently, the commonly used Cartesian coordinates, are sensitive to rigid transformations. To illustrate that, we train \ourmethod{} on the task of semantic segmentation: (i) using invariant geometric features and, (ii) using the edge midpoints $(x,y,z)$ as input features. To assess the learning generalization, we apply non-uniform scaling along the vertical axis (without training on these types of augmentations). Our relative geometric features achieve $98.44\%$, compared with $99.63\%$ on the standard test set, while the absolute coordinates deteriorate to $78.27\%$, compared to $99.11\%$ on the standard test set. Note that while our geometric features are not invariant to non-uniform scaling, they generalize better due to their insensitivity to positioning.}

\section{Discussion and Future work}

We have presented \ourmethod{}, a general method for employing neural networks directly on irregular triangular meshes. The key contribution of our work is the definition and application of convolution and pooling operations tailored to irregular and non-uniform structures. 
These operations facilitate a direct analysis of shapes represented as meshes in their native form, and hence benefit from the unique properties associated with the representation of surface manifolds with non-uniform structures.  

{\bf Invariant Convolutions.} Our choice of mesh edges as basic building blocks upon which the network operates, is of great importance, as the edge set dictates a simple means to define a local, fixed sized neighborhood for convolutions over irregular structures. 
\rh{By exploiting the unique symmetries specific to triangular meshes, we disambiguate the neighbor ordering duality to enable invariance to transformations.} We complement this effort with our choice of input edge features, which are carefully curated to contain solely relative geometric properties rather than absolute positions. Thus, unlike common representations (\emph{e.g.}, point-based), \rhc{I think we can say something stronger here because even the manifold approaches will use points.} the Cartesian coordinates of the vertices are ignored, and local and non-local features are position-oblivious, allowing better generalization of shape features, and facilitating invariance to similarity transformations. We emphasize that we use vertex positions solely for displaying the evolving meshes, but their positions have no impact on the task. 


{\bf Spatially Adaptive Pooling.} We have developed a pooling operation carried out by edge collapse, based on the learned edge features, leading to task-driven pooling guided by the network loss function. \dc{In the future, we would like to add dedicated and separate set of features for prioritizing the edge collapse, similar to attention models.}
Visualizing the series of features the network determined were important, led to profound insights with regards to what the network had actually learned.
We observed that our differential features not only provide invariance to similarity transformations, but also inhibit overfitting, as opposed to the usage of absolute Cartesian coordinates. The generalization of the network was further demonstrated via its ability to perform semantically similar pooling across different objects, naturally leading to better results. Investigating this powerful mechanism may lead to a better understanding of the neural net behavior.

We regard this spatially-adaptive irregular task-driven pooling as an important contribution with the potential to also influence many image-based CNN tasks.  For instance, high resolution image segmentation typically produces a low resolution segmentation map and upsamples it, possibly with skip connections. The pooling in \ourmethod{} semantically simplifies regions with uniform features, while preserving complex ones; therefore, in the future, we are interested in applying similar irregular pooling for image segmentation tasks to obtain high resolution segmentation maps, where large uniform regions in the image will be represented by a small number of triangles.

Currently, our implementation performs sequential edge collapses. This operation can potentially be \rg{parallelized on a GPU by using a parallel sorting technique \cite{Bozidar2015ComparisonOP} for the edge features (calculated only once per pooling operation) and ensuring that only non-adjacent edges are collapsed simultaneously. Clearly, the pooled features in this non-sequential way might be different than in the sequential one}.

Despite the robustness of our method to different triangulations (as demonstrated by our experiments), \ourmethod{}, like any other network, relies on good training data for a successful generalization. In this sense, much like adversarial noise in images, \ourmethod{} is vulnerable to adversarial remeshing attacks that may hinder performance. Robustness to such adversarial attacks is therefore an interesting direction for future work.

Another avenue for future research is generative modeling, mesh-upsampling and attribute synthesis for the modification of existing meshes. Our idea is to apply vertex-split in reverse order of the edge collapse operations by bookkeeping the list of edge collapses. This is similar to the bookkeeping that is used for the unpooling layer. Thus, when synthesizing new meshes, the network decides which vertex to split, for example, by splitting vertices that are adjacent to edges with high feature values.

Finally, we identify a promising endeavor in the extension of our proposed strategy, designed for triangular meshes, to general graphs. Edge-collapse based pooling and unpooling can be applied to general graphs in a similar manner to our proposed \ourmethod{}. As for convolution, we must consider an appropriate alternative suitable for the irregularity of general graphs. An interesting approach may be to process the edges using an attention mechanism~\cite{Monti2018DualPrimalGC}.


\bibliographystyle{ACM-Reference-Format}
\bibliography{bibs} 


\begin{thebibliography}{75}


\ifx \showCODEN    \undefined \def \showCODEN     #1{\unskip}     \fi
\ifx \showDOI      \undefined \def \showDOI       #1{#1}\fi
\ifx \showISBNx    \undefined \def \showISBNx     #1{\unskip}     \fi
\ifx \showISBNxiii \undefined \def \showISBNxiii  #1{\unskip}     \fi
\ifx \showISSN     \undefined \def \showISSN      #1{\unskip}     \fi
\ifx \showLCCN     \undefined \def \showLCCN      #1{\unskip}     \fi
\ifx \shownote     \undefined \def \shownote      #1{#1}          \fi
\ifx \showarticletitle \undefined \def \showarticletitle #1{#1}   \fi
\ifx \showURL      \undefined \def \showURL       {\relax}        \fi
\providecommand\bibfield[2]{#2}
\providecommand\bibinfo[2]{#2}
\providecommand\natexlab[1]{#1}
\providecommand\showeprint[2][]{arXiv:#2}

\bibitem[\protect\citeauthoryear{Adobe}{Adobe}{2016}]%
        {adobe}
\bibfield{author}{\bibinfo{person}{Adobe}.} \bibinfo{year}{2016}\natexlab{}.
\newblock \bibinfo{title}{Adobe Fuse 3D Characters}.
\newblock \bibinfo{howpublished}{\url{https://www.mixamo.com}}.
\newblock


\bibitem[\protect\citeauthoryear{Anguelov, Srinivasan, Koller, Thrun, Rodgers,
  and Davis}{Anguelov et~al\mbox{.}}{2005}]%
        {anguelov2005scape}
\bibfield{author}{\bibinfo{person}{Dragomir Anguelov}, \bibinfo{person}{Praveen
  Srinivasan}, \bibinfo{person}{Daphne Koller}, \bibinfo{person}{Sebastian
  Thrun}, \bibinfo{person}{Jim Rodgers}, {and} \bibinfo{person}{James Davis}.}
  \bibinfo{year}{2005}\natexlab{}.
\newblock \showarticletitle{SCAPE: Shape Completion and Animation of People}.
  In \bibinfo{booktitle}{\emph{ACM SIGGRAPH 2005 Papers}}
  \emph{(\bibinfo{series}{SIGGRAPH '05})}. \bibinfo{publisher}{ACM},
  \bibinfo{address}{New York, NY, USA}, \bibinfo{pages}{408--416}.
\newblock
\urldef\tempurl%
\url{https://doi.org/10.1145/1186822.1073207}
\showDOI{\tempurl}


\bibitem[\protect\citeauthoryear{Atwood and Towsley}{Atwood and
  Towsley}{2016}]%
        {Atwood16Diffusion}
\bibfield{author}{\bibinfo{person}{James Atwood} {and} \bibinfo{person}{Don
  Towsley}.} \bibinfo{year}{2016}\natexlab{}.
\newblock \showarticletitle{Diffusion-convolutional Neural Networks}. In
  \bibinfo{booktitle}{\emph{Proceedings of the 30th International Conference on
  Neural Information Processing Systems}} \emph{(\bibinfo{series}{NIPS'16})}.
  \bibinfo{publisher}{Curran Associates Inc.}, \bibinfo{address}{USA},
  \bibinfo{pages}{2001--2009}.
\newblock
\showISBNx{978-1-5108-3881-9}
\urldef\tempurl%
\url{http://dl.acm.org/citation.cfm?id=3157096.3157320}
\showURL{%
\tempurl}


\bibitem[\protect\citeauthoryear{Atzmon, Maron, and Lipman}{Atzmon
  et~al\mbox{.}}{2018}]%
        {atzmon2018point}
\bibfield{author}{\bibinfo{person}{Matan Atzmon}, \bibinfo{person}{Haggai
  Maron}, {and} \bibinfo{person}{Yaron Lipman}.}
  \bibinfo{year}{2018}\natexlab{}.
\newblock \showarticletitle{Point Convolutional Neural Networks by Extension
  Operators}.
\newblock \bibinfo{journal}{\emph{ACM Trans. Graph.}} \bibinfo{volume}{37},
  \bibinfo{number}{4} (\bibinfo{date}{July} \bibinfo{year}{2018}),
  \bibinfo{pages}{71:1--71:12}.
\newblock


\bibitem[\protect\citeauthoryear{Berg, Cheong, Kreveld, and Overmars}{Berg
  et~al\mbox{.}}{2008}]%
        {Berg:2008}
\bibfield{author}{\bibinfo{person}{Mark~de Berg}, \bibinfo{person}{Otfried
  Cheong}, \bibinfo{person}{Marc~van Kreveld}, {and} \bibinfo{person}{Mark
  Overmars}.} \bibinfo{year}{2008}\natexlab{}.
\newblock \bibinfo{booktitle}{\emph{Computational Geometry: Algorithms and
  Applications} (\bibinfo{edition}{3rd ed.} ed.)}.
\newblock \bibinfo{publisher}{Springer-Verlag TELOS}, \bibinfo{address}{Santa
  Clara, CA, USA}.
\newblock
\showISBNx{3540779736, 9783540779735}


\bibitem[\protect\citeauthoryear{Bogo, Romero, Loper, and Black}{Bogo
  et~al\mbox{.}}{2014}]%
        {bogo2014faust}
\bibfield{author}{\bibinfo{person}{Federica Bogo}, \bibinfo{person}{Javier
  Romero}, \bibinfo{person}{Matthew Loper}, {and} \bibinfo{person}{Michael~J
  Black}.} \bibinfo{year}{2014}\natexlab{}.
\newblock \showarticletitle{FAUST: Dataset and evaluation for 3D mesh
  registration}. In \bibinfo{booktitle}{\emph{Proceedings of the IEEE
  Conference on Computer Vision and Pattern Recognition}}.
  \bibinfo{pages}{3794--3801}.
\newblock


\bibitem[\protect\citeauthoryear{Boscaini, Masci, Melzi, Bronstein, Castellani,
  and Vandergheynst}{Boscaini et~al\mbox{.}}{2015}]%
        {boscaini2015learning}
\bibfield{author}{\bibinfo{person}{Davide Boscaini}, \bibinfo{person}{Jonathan
  Masci}, \bibinfo{person}{Simone Melzi}, \bibinfo{person}{Michael~M
  Bronstein}, \bibinfo{person}{Umberto Castellani}, {and}
  \bibinfo{person}{Pierre Vandergheynst}.} \bibinfo{year}{2015}\natexlab{}.
\newblock \showarticletitle{Learning class-specific descriptors for deformable
  shapes using localized spectral convolutional networks}. In
  \bibinfo{booktitle}{\emph{Computer Graphics Forum}},
  Vol.~\bibinfo{volume}{34}. Wiley Online Library, \bibinfo{pages}{13--23}.
\newblock


\bibitem[\protect\citeauthoryear{Boscaini, Masci, Rodol{\`a}, and
  Bronstein}{Boscaini et~al\mbox{.}}{2016}]%
        {boscaini2016learning}
\bibfield{author}{\bibinfo{person}{Davide Boscaini}, \bibinfo{person}{Jonathan
  Masci}, \bibinfo{person}{Emanuele Rodol{\`a}}, {and} \bibinfo{person}{Michael
  Bronstein}.} \bibinfo{year}{2016}\natexlab{}.
\newblock \showarticletitle{Learning shape correspondence with anisotropic
  convolutional neural networks}. In \bibinfo{booktitle}{\emph{Advances in
  Neural Information Processing Systems}}. \bibinfo{pages}{3189--3197}.
\newblock


\bibitem[\protect\citeauthoryear{Botsch, Kobbelt, Pauly, Alliez, and
  L{\'e}vy}{Botsch et~al\mbox{.}}{2010}]%
        {botsch2010polygon}
\bibfield{author}{\bibinfo{person}{Mario Botsch}, \bibinfo{person}{Leif
  Kobbelt}, \bibinfo{person}{Mark Pauly}, \bibinfo{person}{Pierre Alliez},
  {and} \bibinfo{person}{Bruno L{\'e}vy}.} \bibinfo{year}{2010}\natexlab{}.
\newblock \bibinfo{booktitle}{\emph{Polygon mesh processing}}.
\newblock \bibinfo{publisher}{AK Peters/CRC Press}.
\newblock


\bibitem[\protect\citeauthoryear{Bozidar and Dobravec}{Bozidar and
  Dobravec}{2015}]%
        {Bozidar2015ComparisonOP}
\bibfield{author}{\bibinfo{person}{Darko Bozidar} {and} \bibinfo{person}{Tomaz
  Dobravec}.} \bibinfo{year}{2015}\natexlab{}.
\newblock \showarticletitle{Comparison of parallel sorting algorithms}.
\newblock \bibinfo{journal}{\emph{CoRR}}  \bibinfo{volume}{abs/1511.03404}
  (\bibinfo{year}{2015}).
\newblock


\bibitem[\protect\citeauthoryear{Brock, Lim, Ritchie, and Weston}{Brock
  et~al\mbox{.}}{2016}]%
        {Brock16Generative}
\bibfield{author}{\bibinfo{person}{Andrew Brock}, \bibinfo{person}{Theodore
  Lim}, \bibinfo{person}{J.M. Ritchie}, {and} \bibinfo{person}{Nick Weston}.}
  \bibinfo{year}{2016}\natexlab{}.
\newblock \showarticletitle{Generative and Discriminative Voxel Modeling with
  Convolutional Neural Networks}. In \bibinfo{booktitle}{\emph{NIPS 3D Deep
  Learning Workshop}}.
\newblock


\bibitem[\protect\citeauthoryear{Bronstein, Bronstein, Guibas, and
  Ovsjanikov}{Bronstein et~al\mbox{.}}{2011}]%
        {bronstein2011shape}
\bibfield{author}{\bibinfo{person}{Alexander~M Bronstein},
  \bibinfo{person}{Michael~M Bronstein}, \bibinfo{person}{Leonidas~J Guibas},
  {and} \bibinfo{person}{Maks Ovsjanikov}.} \bibinfo{year}{2011}\natexlab{}.
\newblock \showarticletitle{Shape google: Geometric words and expressions for
  invariant shape retrieval}.
\newblock \bibinfo{journal}{\emph{ACM Transactions on Graphics (TOG)}}
  \bibinfo{volume}{30}, \bibinfo{number}{1} (\bibinfo{year}{2011}),
  \bibinfo{pages}{1}.
\newblock


\bibitem[\protect\citeauthoryear{Bronstein, Bruna, LeCun, Szlam, and
  Vandergheynst}{Bronstein et~al\mbox{.}}{2017}]%
        {Bronstein17Geometric}
\bibfield{author}{\bibinfo{person}{Michael~M. Bronstein}, \bibinfo{person}{Joan
  Bruna}, \bibinfo{person}{Yann LeCun}, \bibinfo{person}{Arthur Szlam}, {and}
  \bibinfo{person}{Pierre Vandergheynst}.} \bibinfo{year}{2017}\natexlab{}.
\newblock \showarticletitle{Geometric Deep Learning: Going beyond Euclidean
  data}.
\newblock \bibinfo{journal}{\emph{{IEEE} Signal Process. Mag.}}
  \bibinfo{volume}{34}, \bibinfo{number}{4} (\bibinfo{year}{2017}),
  \bibinfo{pages}{18--42}.
\newblock
\urldef\tempurl%
\url{https://doi.org/10.1109/MSP.2017.2693418}
\showDOI{\tempurl}


\bibitem[\protect\citeauthoryear{Bruna, Zaremba, Szlam, and LeCun}{Bruna
  et~al\mbox{.}}{2014}]%
        {Bruna14Spectral}
\bibfield{author}{\bibinfo{person}{Joan Bruna}, \bibinfo{person}{Wojciech
  Zaremba}, \bibinfo{person}{Arthur Szlam}, {and} \bibinfo{person}{Yann
  LeCun}.} \bibinfo{year}{2014}\natexlab{}.
\newblock \showarticletitle{Spectral Networks and Locally Connected Networks on
  Graphs}. In \bibinfo{booktitle}{\emph{International Conference on Learning
  Representations (ICLR)}}.
\newblock


\bibitem[\protect\citeauthoryear{Cangea, Velickovic, Jovanovic, Kipf, and
  Lio}{Cangea et~al\mbox{.}}{2018}]%
        {Cangea18Hierarchical}
\bibfield{author}{\bibinfo{person}{C. Cangea}, \bibinfo{person}{P. Velickovic},
  \bibinfo{person}{N. Jovanovic}, \bibinfo{person}{T. Kipf}, {and}
  \bibinfo{person}{P. Lio}.} \bibinfo{year}{2018}\natexlab{}.
\newblock \showarticletitle{Towards Sparse Hierarchical Graph Classifiers}. In
  \bibinfo{booktitle}{\emph{NeurIPS Workshop on Relational Representation
  Learning}}.
\newblock


\bibitem[\protect\citeauthoryear{Chen, Papandreou, Kokkinos, Murphy, and
  Yuille}{Chen et~al\mbox{.}}{2018}]%
        {chen2018deeplab}
\bibfield{author}{\bibinfo{person}{Liang-Chieh Chen}, \bibinfo{person}{George
  Papandreou}, \bibinfo{person}{Iasonas Kokkinos}, \bibinfo{person}{Kevin
  Murphy}, {and} \bibinfo{person}{Alan~L Yuille}.}
  \bibinfo{year}{2018}\natexlab{}.
\newblock \showarticletitle{Deeplab: Semantic image segmentation with deep
  convolutional nets, atrous convolution, and fully connected crfs}.
\newblock \bibinfo{journal}{\emph{IEEE transactions on pattern analysis and
  machine intelligence}} \bibinfo{volume}{40}, \bibinfo{number}{4}
  (\bibinfo{year}{2018}), \bibinfo{pages}{834--848}.
\newblock


\bibitem[\protect\citeauthoryear{Defferrard, Bresson, and
  Vandergheynst}{Defferrard et~al\mbox{.}}{2016}]%
        {defferrard2016convolutional}
\bibfield{author}{\bibinfo{person}{Micha{\"e}l Defferrard},
  \bibinfo{person}{Xavier Bresson}, {and} \bibinfo{person}{Pierre
  Vandergheynst}.} \bibinfo{year}{2016}\natexlab{}.
\newblock \showarticletitle{Convolutional neural networks on graphs with fast
  localized spectral filtering}. In \bibinfo{booktitle}{\emph{Advances in
  Neural Information Processing Systems}}. \bibinfo{pages}{3844--3852}.
\newblock


\bibitem[\protect\citeauthoryear{Ezuz, Solomon, Kim, and Ben-Chen}{Ezuz
  et~al\mbox{.}}{2017}]%
        {gwcnn}
\bibfield{author}{\bibinfo{person}{Danielle Ezuz}, \bibinfo{person}{Justin
  Solomon}, \bibinfo{person}{Vladimir~G. Kim}, {and} \bibinfo{person}{Mirela
  Ben-Chen}.} \bibinfo{year}{2017}\natexlab{}.
\newblock \showarticletitle{{GWCNN: A Metric Alignment Layer for Deep Shape
  Analysis}}.
\newblock \bibinfo{journal}{\emph{Computer Graphics Forum}}
  (\bibinfo{year}{2017}).
\newblock
\showISSN{1467-8659}
\urldef\tempurl%
\url{https://doi.org/10.1111/cgf.13244}
\showDOI{\tempurl}


\bibitem[\protect\citeauthoryear{Gao, Panozzo, Wang, Deng, and Chen}{Gao
  et~al\mbox{.}}{2017}]%
        {gao2017robust}
\bibfield{author}{\bibinfo{person}{Xifeng Gao}, \bibinfo{person}{Daniele
  Panozzo}, \bibinfo{person}{Wenping Wang}, \bibinfo{person}{Zhigang Deng},
  {and} \bibinfo{person}{Guoning Chen}.} \bibinfo{year}{2017}\natexlab{}.
\newblock \showarticletitle{Robust structure simplification for hex
  re-meshing}.
\newblock \bibinfo{journal}{\emph{ACM Transactions on Graphics}}
  \bibinfo{volume}{36}, \bibinfo{number}{6} (\bibinfo{year}{2017}).
\newblock


\bibitem[\protect\citeauthoryear{Garland and Heckbert}{Garland and
  Heckbert}{1997}]%
        {garland1997surface}
\bibfield{author}{\bibinfo{person}{Michael Garland} {and}
  \bibinfo{person}{Paul~S Heckbert}.} \bibinfo{year}{1997}\natexlab{}.
\newblock \showarticletitle{Surface simplification using quadric error
  metrics}. In \bibinfo{booktitle}{\emph{Proceedings of the 24th annual
  conference on Computer graphics and interactive techniques}}. ACM
  Press/Addison-Wesley Publishing Co., \bibinfo{pages}{209--216}.
\newblock


\bibitem[\protect\citeauthoryear{Giorgi, Biasotti, and Paraboschi}{Giorgi
  et~al\mbox{.}}{2007}]%
        {giorgi2007shape}
\bibfield{author}{\bibinfo{person}{Daniela Giorgi}, \bibinfo{person}{Silvia
  Biasotti}, {and} \bibinfo{person}{Laura Paraboschi}.}
  \bibinfo{year}{2007}\natexlab{}.
\newblock \showarticletitle{Shape retrieval contest 2007: Watertight models
  track}.
\newblock \bibinfo{journal}{\emph{SHREC competition}} \bibinfo{volume}{8},
  \bibinfo{number}{7} (\bibinfo{year}{2007}).
\newblock


\bibitem[\protect\citeauthoryear{Graham, Engelcke, and van~der Maaten}{Graham
  et~al\mbox{.}}{2017}]%
        {Graham17Semantic}
\bibfield{author}{\bibinfo{person}{Benjamin Graham}, \bibinfo{person}{Martin
  Engelcke}, {and} \bibinfo{person}{Laurens van~der Maaten}.}
  \bibinfo{year}{2017}\natexlab{}.
\newblock \showarticletitle{{3D} Semantic Segmentation with Submanifold Sparse
  Convolutional Networks}.
\newblock \bibinfo{journal}{\emph{CoRR}}  \bibinfo{volume}{abs/1711.10275}
  (\bibinfo{year}{2017}).
\newblock


\bibitem[\protect\citeauthoryear{Guerrero, Kleiman, Ovsjanikov, and
  Mitra}{Guerrero et~al\mbox{.}}{2018}]%
        {GuerreroEtAl:PCPNet:EG:2018}
\bibfield{author}{\bibinfo{person}{Paul Guerrero}, \bibinfo{person}{Yanir
  Kleiman}, \bibinfo{person}{Maks Ovsjanikov}, {and} \bibinfo{person}{Niloy~J.
  Mitra}.} \bibinfo{year}{2018}\natexlab{}.
\newblock \showarticletitle{{PCPNet}: Learning Local Shape Properties from Raw
  Point Clouds}.
\newblock \bibinfo{journal}{\emph{Computer Graphics Forum}}
  \bibinfo{volume}{37}, \bibinfo{number}{2} (\bibinfo{year}{2018}),
  \bibinfo{pages}{75--85}.
\newblock
\urldef\tempurl%
\url{https://doi.org/10.1111/cgf.13343}
\showDOI{\tempurl}


\bibitem[\protect\citeauthoryear{Haim, Segol, Ben-Hamu, Maron, and Lipman}{Haim
  et~al\mbox{.}}{2018}]%
        {Haim2018Surface}
\bibfield{author}{\bibinfo{person}{Niv Haim}, \bibinfo{person}{Nimrod Segol},
  \bibinfo{person}{Heli Ben-Hamu}, \bibinfo{person}{Haggai Maron}, {and}
  \bibinfo{person}{Yaron Lipman}.} \bibinfo{year}{2018}\natexlab{}.
\newblock \showarticletitle{Surface Networks via General Covers}.
\newblock \bibinfo{journal}{\emph{CoRR}}  \bibinfo{volume}{abs/1812.10705}
  (\bibinfo{year}{2018}).
\newblock


\bibitem[\protect\citeauthoryear{Hanocka, Fish, Wang, Giryes, Fleishman, and
  Cohen-Or}{Hanocka et~al\mbox{.}}{2018}]%
        {hanocka2018alignet}
\bibfield{author}{\bibinfo{person}{Rana Hanocka}, \bibinfo{person}{Noa Fish},
  \bibinfo{person}{Zhenhua Wang}, \bibinfo{person}{Raja Giryes},
  \bibinfo{person}{Shachar Fleishman}, {and} \bibinfo{person}{Daniel
  Cohen-Or}.} \bibinfo{year}{2018}\natexlab{}.
\newblock \showarticletitle{ALIGNet: Partial-Shape Agnostic Alignment via
  Unsupervised Learning}.
\newblock \bibinfo{journal}{\emph{ACM Trans. Graph.}} \bibinfo{volume}{38},
  \bibinfo{number}{1}, Article \bibinfo{articleno}{1} (\bibinfo{date}{Dec.}
  \bibinfo{year}{2018}), \bibinfo{numpages}{14}~pages.
\newblock
\showISSN{0730-0301}
\urldef\tempurl%
\url{https://doi.org/10.1145/3267347}
\showDOI{\tempurl}


\bibitem[\protect\citeauthoryear{Henaff, Bruna, and LeCun}{Henaff
  et~al\mbox{.}}{2015}]%
        {Henaff15Deep}
\bibfield{author}{\bibinfo{person}{Mikael Henaff}, \bibinfo{person}{Joan
  Bruna}, {and} \bibinfo{person}{Yann LeCun}.} \bibinfo{year}{2015}\natexlab{}.
\newblock \showarticletitle{Deep Convolutional Networks on Graph-Structured
  Data}.
\newblock \bibinfo{journal}{\emph{CoRR}}  \bibinfo{volume}{abs/1506.05163}
  (\bibinfo{year}{2015}).
\newblock


\bibitem[\protect\citeauthoryear{Hoppe}{Hoppe}{1997}]%
        {hoppe1997view}
\bibfield{author}{\bibinfo{person}{Hugues Hoppe}.}
  \bibinfo{year}{1997}\natexlab{}.
\newblock \showarticletitle{View-dependent refinement of progressive meshes}.
  In \bibinfo{booktitle}{\emph{Proceedings of the 24th annual conference on
  Computer graphics and interactive techniques}}. ACM Press/Addison-Wesley
  Publishing Co., \bibinfo{pages}{189--198}.
\newblock


\bibitem[\protect\citeauthoryear{Hoppe}{Hoppe}{1999}]%
        {hoppe1999new}
\bibfield{author}{\bibinfo{person}{Hugues Hoppe}.}
  \bibinfo{year}{1999}\natexlab{}.
\newblock \showarticletitle{New quadric metric for simplifying meshes with
  appearance attributes}. In \bibinfo{booktitle}{\emph{Visualization'99.
  Proceedings}}. IEEE, \bibinfo{pages}{59--510}.
\newblock


\bibitem[\protect\citeauthoryear{Hoppe, DeRose, Duchamp, McDonald, and
  Stuetzle}{Hoppe et~al\mbox{.}}{1993}]%
        {hoppe1993mesh}
\bibfield{author}{\bibinfo{person}{Hugues Hoppe}, \bibinfo{person}{Tony
  DeRose}, \bibinfo{person}{Tom Duchamp}, \bibinfo{person}{John McDonald},
  {and} \bibinfo{person}{Werner Stuetzle}.} \bibinfo{year}{1993}\natexlab{}.
\newblock \bibinfo{title}{Mesh optimization}.
\newblock , \bibinfo{numpages}{19--26}~pages.
\newblock


\bibitem[\protect\citeauthoryear{Jia}{Jia}{2014}]%
        {jia2014learning}
\bibfield{author}{\bibinfo{person}{Yangqing Jia}.}
  \bibinfo{year}{2014}\natexlab{}.
\newblock \showarticletitle{Learning Semantic Image Representations at a Large
  Scale}.
\newblock  (\bibinfo{year}{2014}).
\newblock


\bibitem[\protect\citeauthoryear{Kalogerakis, Averkiou, Maji, and
  Chaudhuri}{Kalogerakis et~al\mbox{.}}{2017}]%
        {kalogerakis20173d}
\bibfield{author}{\bibinfo{person}{Evangelos Kalogerakis},
  \bibinfo{person}{Melinos Averkiou}, \bibinfo{person}{Subhransu Maji}, {and}
  \bibinfo{person}{Siddhartha Chaudhuri}.} \bibinfo{year}{2017}\natexlab{}.
\newblock \showarticletitle{3D shape segmentation with projective convolutional
  networks}. In \bibinfo{booktitle}{\emph{Proc. CVPR}},
  Vol.~\bibinfo{volume}{1}. \bibinfo{pages}{8}.
\newblock


\bibitem[\protect\citeauthoryear{Kalogerakis, Hertzmann, and Singh}{Kalogerakis
  et~al\mbox{.}}{2010}]%
        {kalogerakis2010learning}
\bibfield{author}{\bibinfo{person}{Evangelos Kalogerakis},
  \bibinfo{person}{Aaron Hertzmann}, {and} \bibinfo{person}{Karan Singh}.}
  \bibinfo{year}{2010}\natexlab{}.
\newblock \showarticletitle{Learning 3D mesh segmentation and labeling}.
\newblock \bibinfo{journal}{\emph{ACM Transactions on Graphics (TOG)}}
  \bibinfo{volume}{29}, \bibinfo{number}{4} (\bibinfo{year}{2010}),
  \bibinfo{pages}{102}.
\newblock


\bibitem[\protect\citeauthoryear{Kokkinos, Bronstein, Litman, and
  Bronstein}{Kokkinos et~al\mbox{.}}{2012}]%
        {Kokkinos12Intrinsic}
\bibfield{author}{\bibinfo{person}{I. Kokkinos}, \bibinfo{person}{M.~M.
  Bronstein}, \bibinfo{person}{R. Litman}, {and} \bibinfo{person}{A.~M.
  Bronstein}.} \bibinfo{year}{2012}\natexlab{}.
\newblock \showarticletitle{Intrinsic shape context descriptors for deformable
  shapes}. In \bibinfo{booktitle}{\emph{IEEE Conference on Computer Vision and
  Pattern Recognition (CVPR)}}. \bibinfo{pages}{159--166}.
\newblock


\bibitem[\protect\citeauthoryear{Kostrikov, Jiang, Panozzo, Zorin, and
  Joan}{Kostrikov et~al\mbox{.}}{2018}]%
        {kostrikov2018surface}
\bibfield{author}{\bibinfo{person}{Ilya Kostrikov}, \bibinfo{person}{Zhongshi
  Jiang}, \bibinfo{person}{Daniele Panozzo}, \bibinfo{person}{Denis Zorin},
  {and} \bibinfo{person}{Burna Joan}.} \bibinfo{year}{2018}\natexlab{}.
\newblock \showarticletitle{Surface Networks}. In
  \bibinfo{booktitle}{\emph{{IEEE} Conference on Computer Vision and Pattern
  Recognition ({CVPR})}}.
\newblock


\bibitem[\protect\citeauthoryear{Krizhevsky, Sutskever, and Hinton}{Krizhevsky
  et~al\mbox{.}}{2012}]%
        {krizhevsky2012imagenet}
\bibfield{author}{\bibinfo{person}{Alex Krizhevsky}, \bibinfo{person}{Ilya
  Sutskever}, {and} \bibinfo{person}{Geoffrey~E Hinton}.}
  \bibinfo{year}{2012}\natexlab{}.
\newblock \showarticletitle{Imagenet classification with deep convolutional
  neural networks}. In \bibinfo{booktitle}{\emph{Advances in neural information
  processing systems}}. \bibinfo{pages}{1097--1105}.
\newblock


\bibitem[\protect\citeauthoryear{Latecki and Lakamper}{Latecki and
  Lakamper}{2000}]%
        {latecki2000shape}
\bibfield{author}{\bibinfo{person}{Longin~Jan Latecki} {and}
  \bibinfo{person}{Rolf Lakamper}.} \bibinfo{year}{2000}\natexlab{}.
\newblock \showarticletitle{Shape similarity measure based on correspondence of
  visual parts}.
\newblock \bibinfo{journal}{\emph{IEEE Transactions on Pattern Analysis and
  Machine Intelligence}} \bibinfo{volume}{22}, \bibinfo{number}{10}
  (\bibinfo{year}{2000}), \bibinfo{pages}{1185--1190}.
\newblock


\bibitem[\protect\citeauthoryear{LeCun}{LeCun}{2012}]%
        {lecun2012learning}
\bibfield{author}{\bibinfo{person}{Yann LeCun}.}
  \bibinfo{year}{2012}\natexlab{}.
\newblock \showarticletitle{Learning invariant feature hierarchies}. In
  \bibinfo{booktitle}{\emph{European conference on computer vision}}. Springer,
  \bibinfo{pages}{496--505}.
\newblock


\bibitem[\protect\citeauthoryear{Li, Bu, Sun, and Chen}{Li
  et~al\mbox{.}}{2018}]%
        {Li18PointCNN}
\bibfield{author}{\bibinfo{person}{Yangyan Li}, \bibinfo{person}{Rui Bu},
  \bibinfo{person}{Mingchao Sun}, {and} \bibinfo{person}{Baoquan Chen}.}
  \bibinfo{year}{2018}\natexlab{}.
\newblock \showarticletitle{PointCNN}.
\newblock \bibinfo{journal}{\emph{CoRR}}  \bibinfo{volume}{abs/1801.07791}
  (\bibinfo{year}{2018}).
\newblock


\bibitem[\protect\citeauthoryear{Li, Pirk, Su, Qi, and Guibas}{Li
  et~al\mbox{.}}{2016}]%
        {Li16FPNN}
\bibfield{author}{\bibinfo{person}{Yangyan Li}, \bibinfo{person}{Soren Pirk},
  \bibinfo{person}{Hao Su}, \bibinfo{person}{Charles~R Qi}, {and}
  \bibinfo{person}{Leonidas~J Guibas}.} \bibinfo{year}{2016}\natexlab{}.
\newblock \showarticletitle{FPNN: Field probing neural networks for {3D} data}.
  In \bibinfo{booktitle}{\emph{Advances in Neural Information Processing
  Systems (NIPS)}}. \bibinfo{pages}{307–315}.
\newblock


\bibitem[\protect\citeauthoryear{Lian, Godil, Bustos, Daoudi, Hermans,
  Kawamura, Kurita, Lavoua, and Dp~Suetens}{Lian et~al\mbox{.}}{2011}]%
        {shrec2011}
\bibfield{author}{\bibinfo{person}{Z Lian}, \bibinfo{person}{A Godil},
  \bibinfo{person}{B Bustos}, \bibinfo{person}{M Daoudi}, \bibinfo{person}{J
  Hermans}, \bibinfo{person}{S Kawamura}, \bibinfo{person}{Y Kurita},
  \bibinfo{person}{G Lavoua}, {and} \bibinfo{person}{P Dp~Suetens}.}
  \bibinfo{year}{2011}\natexlab{}.
\newblock \showarticletitle{Shape retrieval on non-rigid 3D watertight meshes}.
  In \bibinfo{booktitle}{\emph{Eurographics Workshop on 3D Object Retrieval
  (3DOR)}}.
\newblock


\bibitem[\protect\citeauthoryear{Litany, Bronstein, Bronstein, and
  Makadia}{Litany et~al\mbox{.}}{2018}]%
        {Litany2018DeformableSC}
\bibfield{author}{\bibinfo{person}{Or Litany}, \bibinfo{person}{Alexander~M.
  Bronstein}, \bibinfo{person}{Michael~M. Bronstein}, {and}
  \bibinfo{person}{Ameesh Makadia}.} \bibinfo{year}{2018}\natexlab{}.
\newblock \showarticletitle{Deformable Shape Completion With Graph
  Convolutional Autoencoders}. In \bibinfo{booktitle}{\emph{CVPR}}.
\newblock


\bibitem[\protect\citeauthoryear{Maron, Galun, Aigerman, Trope, Dym, Yumer,
  Kim, and Lipman}{Maron et~al\mbox{.}}{2017}]%
        {maron2017convolutional}
\bibfield{author}{\bibinfo{person}{Haggai Maron}, \bibinfo{person}{Meirav
  Galun}, \bibinfo{person}{Noam Aigerman}, \bibinfo{person}{Miri Trope},
  \bibinfo{person}{Nadav Dym}, \bibinfo{person}{Ersin Yumer},
  \bibinfo{person}{Vladimir~G Kim}, {and} \bibinfo{person}{Yaron Lipman}.}
  \bibinfo{year}{2017}\natexlab{}.
\newblock \showarticletitle{Convolutional neural networks on surfaces via
  seamless toric covers}.
\newblock \bibinfo{journal}{\emph{ACM Trans. Graph}} \bibinfo{volume}{36},
  \bibinfo{number}{4} (\bibinfo{year}{2017}), \bibinfo{pages}{71}.
\newblock


\bibitem[\protect\citeauthoryear{Masci, Boscaini, Bronstein, and
  Vandergheynst}{Masci et~al\mbox{.}}{2015}]%
        {Masci15Geodesic}
\bibfield{author}{\bibinfo{person}{Jonathan Masci}, \bibinfo{person}{Davide
  Boscaini}, \bibinfo{person}{Michael Bronstein}, {and} \bibinfo{person}{Pierre
  Vandergheynst}.} \bibinfo{year}{2015}\natexlab{}.
\newblock \showarticletitle{Geodesic convolutional neural networks on
  riemannian manifolds}. In \bibinfo{booktitle}{\emph{Proceedings of the IEEE
  international conference on computer vision workshops}}.
  \bibinfo{pages}{37--45}.
\newblock


\bibitem[\protect\citeauthoryear{Monti, Boscaini, Masci, Rodola, Svoboda, and
  Bronstein}{Monti et~al\mbox{.}}{2017}]%
        {monti2017geometric}
\bibfield{author}{\bibinfo{person}{Federico Monti}, \bibinfo{person}{Davide
  Boscaini}, \bibinfo{person}{Jonathan Masci}, \bibinfo{person}{Emanuele
  Rodola}, \bibinfo{person}{Jan Svoboda}, {and} \bibinfo{person}{Michael~M
  Bronstein}.} \bibinfo{year}{2017}\natexlab{}.
\newblock \showarticletitle{Geometric deep learning on graphs and manifolds
  using mixture model CNNs}. In \bibinfo{booktitle}{\emph{Proc. CVPR}},
  Vol.~\bibinfo{volume}{1}. \bibinfo{pages}{3}.
\newblock


\bibitem[\protect\citeauthoryear{Monti, Shchur, Bojchevski, Litany, Gunnemann,
  and Bronstein}{Monti et~al\mbox{.}}{2018}]%
        {Monti2018DualPrimalGC}
\bibfield{author}{\bibinfo{person}{Federico Monti}, \bibinfo{person}{Oleksandr
  Shchur}, \bibinfo{person}{Aleksandar Bojchevski}, \bibinfo{person}{Or
  Litany}, \bibinfo{person}{Stephan Gunnemann}, {and}
  \bibinfo{person}{Michael~M. Bronstein}.} \bibinfo{year}{2018}\natexlab{}.
\newblock \showarticletitle{Dual-Primal Graph Convolutional Networks}.
\newblock \bibinfo{journal}{\emph{CoRR}}  \bibinfo{volume}{abs/1806.00770}
  (\bibinfo{year}{2018}).
\newblock


\bibitem[\protect\citeauthoryear{Niepert, Ahmed, and Kutzkov}{Niepert
  et~al\mbox{.}}{2016}]%
        {Niepert16Learning}
\bibfield{author}{\bibinfo{person}{Mathias Niepert}, \bibinfo{person}{Mohamed
  Ahmed}, {and} \bibinfo{person}{Konstantin Kutzkov}.}
  \bibinfo{year}{2016}\natexlab{}.
\newblock \showarticletitle{Learning Convolutional Neural Networks for Graphs}.
  In \bibinfo{booktitle}{\emph{International Conference on Machine Learning
  (ICML)}}.
\newblock


\bibitem[\protect\citeauthoryear{Paszke, Gross, Chintala, Chanan, Yang, DeVito,
  Lin, Desmaison, Antiga, and Lerer}{Paszke et~al\mbox{.}}{2017}]%
        {paszke2017automatic}
\bibfield{author}{\bibinfo{person}{Adam Paszke}, \bibinfo{person}{Sam Gross},
  \bibinfo{person}{Soumith Chintala}, \bibinfo{person}{Gregory Chanan},
  \bibinfo{person}{Edward Yang}, \bibinfo{person}{Zachary DeVito},
  \bibinfo{person}{Zeming Lin}, \bibinfo{person}{Alban Desmaison},
  \bibinfo{person}{Luca Antiga}, {and} \bibinfo{person}{Adam Lerer}.}
  \bibinfo{year}{2017}\natexlab{}.
\newblock \showarticletitle{Automatic differentiation in PyTorch}. In
  \bibinfo{booktitle}{\emph{NIPS-W}}.
\newblock


\bibitem[\protect\citeauthoryear{Poulenard and Ovsjanikov}{Poulenard and
  Ovsjanikov}{2018}]%
        {Poulenard}
\bibfield{author}{\bibinfo{person}{Adrien Poulenard} {and}
  \bibinfo{person}{Maks Ovsjanikov}.} \bibinfo{year}{2018}\natexlab{}.
\newblock \showarticletitle{Multi-directional Geodesic Neural Networks via
  Equivariant Convolution}. In \bibinfo{booktitle}{\emph{SIGGRAPH Asia 2018
  Technical Papers}} \emph{(\bibinfo{series}{SIGGRAPH Asia '18})}.
  \bibinfo{publisher}{ACM}, \bibinfo{address}{New York, NY, USA}, Article
  \bibinfo{articleno}{236}, \bibinfo{numpages}{14}~pages.
\newblock
\showISBNx{978-1-4503-6008-1}
\urldef\tempurl%
\url{https://doi.org/10.1145/3272127.3275102}
\showDOI{\tempurl}


\bibitem[\protect\citeauthoryear{Qi, Su, Mo, and Guibas}{Qi
  et~al\mbox{.}}{2017a}]%
        {qi2017pointnet}
\bibfield{author}{\bibinfo{person}{Charles~R Qi}, \bibinfo{person}{Hao Su},
  \bibinfo{person}{Kaichun Mo}, {and} \bibinfo{person}{Leonidas~J Guibas}.}
  \bibinfo{year}{2017}\natexlab{a}.
\newblock \showarticletitle{Pointnet: Deep learning on point sets for 3d
  classification and segmentation}.
\newblock \bibinfo{journal}{\emph{Proc. Computer Vision and Pattern Recognition
  (CVPR), IEEE}} \bibinfo{volume}{1}, \bibinfo{number}{2}
  (\bibinfo{year}{2017}), \bibinfo{pages}{4}.
\newblock


\bibitem[\protect\citeauthoryear{Qi, Su, Niessner, Dai, Yan, and Guibas}{Qi
  et~al\mbox{.}}{2016}]%
        {Qi16Volumetric}
\bibfield{author}{\bibinfo{person}{Charles~R. Qi}, \bibinfo{person}{Hao Su},
  \bibinfo{person}{Matthias Niessner}, \bibinfo{person}{Angela Dai},
  \bibinfo{person}{Mengyuan Yan}, {and} \bibinfo{person}{Leonidas~J. Guibas}.}
  \bibinfo{year}{2016}\natexlab{}.
\newblock \showarticletitle{Volumetric and multi-view CNNs for object
  classification on 3d data}. In \bibinfo{booktitle}{\emph{Computer Vision and
  Pattern Recognition (CVPR)}}. \bibinfo{pages}{5648–5656}.
\newblock


\bibitem[\protect\citeauthoryear{Qi, Yi, Su, and Guibas}{Qi
  et~al\mbox{.}}{2017b}]%
        {Qi17PointNetpp}
\bibfield{author}{\bibinfo{person}{Charles~R. Qi}, \bibinfo{person}{Li Yi},
  \bibinfo{person}{Hao Su}, {and} \bibinfo{person}{Leonidas~J Guibas}.}
  \bibinfo{year}{2017}\natexlab{b}.
\newblock \showarticletitle{PointNet++: Deep Hierarchical Feature Learning on
  Point Sets in a Metric Space}. In \bibinfo{booktitle}{\emph{Advances in
  Neural Information Processing Systems (NIPS)}}. \bibinfo{pages}{5105–5114}.
\newblock


\bibitem[\protect\citeauthoryear{Ranjan, Bolkart, Sanyal, and Black}{Ranjan
  et~al\mbox{.}}{2018}]%
        {Ranjan18Generating}
\bibfield{author}{\bibinfo{person}{Anurag Ranjan}, \bibinfo{person}{Timo
  Bolkart}, \bibinfo{person}{Soubhik Sanyal}, {and} \bibinfo{person}{Michael~J.
  Black}.} \bibinfo{year}{2018}\natexlab{}.
\newblock \showarticletitle{Generating {3D} faces using Convolutional Mesh
  Autoencoders}. In \bibinfo{booktitle}{\emph{European Conference on Computer
  Vision (ECCV)}}. \bibinfo{publisher}{Springer International Publishing},
  \bibinfo{pages}{725--741}.
\newblock


\bibitem[\protect\citeauthoryear{Riegler, Ulusoy, and Geiger}{Riegler
  et~al\mbox{.}}{2017}]%
        {Riegler17OctNet}
\bibfield{author}{\bibinfo{person}{Gernot Riegler}, \bibinfo{person}{Ali~Osman
  Ulusoy}, {and} \bibinfo{person}{Andreas Geiger}.}
  \bibinfo{year}{2017}\natexlab{}.
\newblock \showarticletitle{OctNet: Learning deep 3D representations at high
  resolutions}. In \bibinfo{booktitle}{\emph{Computer Vision and Pattern
  Recognition (CVPR)}}.
\newblock


\bibitem[\protect\citeauthoryear{Ronneberger, Fischer, and Brox}{Ronneberger
  et~al\mbox{.}}{2015}]%
        {ronneberger2015u}
\bibfield{author}{\bibinfo{person}{Olaf Ronneberger}, \bibinfo{person}{Philipp
  Fischer}, {and} \bibinfo{person}{Thomas Brox}.}
  \bibinfo{year}{2015}\natexlab{}.
\newblock \showarticletitle{U-net: Convolutional networks for biomedical image
  segmentation}. In \bibinfo{booktitle}{\emph{International Conference on
  Medical image computing and computer-assisted intervention}}. Springer,
  \bibinfo{pages}{234--241}.
\newblock


\bibitem[\protect\citeauthoryear{Rusinkiewicz and Levoy}{Rusinkiewicz and
  Levoy}{2000}]%
        {Rusinkiewicz:2000:QMP:344779.344940}
\bibfield{author}{\bibinfo{person}{Szymon Rusinkiewicz} {and}
  \bibinfo{person}{Marc Levoy}.} \bibinfo{year}{2000}\natexlab{}.
\newblock \showarticletitle{QSplat: A Multiresolution Point Rendering System
  for Large Meshes}. In \bibinfo{booktitle}{\emph{Proceedings of the 27th
  Annual Conference on Computer Graphics and Interactive Techniques}}
  \emph{(\bibinfo{series}{SIGGRAPH '00})}. \bibinfo{publisher}{ACM
  Press/Addison-Wesley Publishing Co.}, \bibinfo{address}{New York, NY, USA},
  \bibinfo{pages}{343--352}.
\newblock
\showISBNx{1-58113-208-5}
\urldef\tempurl%
\url{https://doi.org/10.1145/344779.344940}
\showDOI{\tempurl}


\bibitem[\protect\citeauthoryear{Sermanet, Eigen, Zhang, Mathieu, Fergus, and
  LeCun}{Sermanet et~al\mbox{.}}{2013}]%
        {sermanet2013overfeat}
\bibfield{author}{\bibinfo{person}{Pierre Sermanet}, \bibinfo{person}{David
  Eigen}, \bibinfo{person}{Xiang Zhang}, \bibinfo{person}{Micha{\"e}l Mathieu},
  \bibinfo{person}{Rob Fergus}, {and} \bibinfo{person}{Yann LeCun}.}
  \bibinfo{year}{2013}\natexlab{}.
\newblock \showarticletitle{Overfeat: Integrated recognition, localization and
  detection using convolutional networks}.
\newblock \bibinfo{journal}{\emph{arXiv preprint arXiv:1312.6229}}
  (\bibinfo{year}{2013}).
\newblock


\bibitem[\protect\citeauthoryear{Simonyan and Zisserman}{Simonyan and
  Zisserman}{2014}]%
        {vgg}
\bibfield{author}{\bibinfo{person}{Karen Simonyan} {and}
  \bibinfo{person}{Andrew Zisserman}.} \bibinfo{year}{2014}\natexlab{}.
\newblock \showarticletitle{Very deep convolutional networks for large-scale
  image recognition}.
\newblock \bibinfo{journal}{\emph{arXiv preprint arXiv:1409.1556}}
  (\bibinfo{year}{2014}).
\newblock


\bibitem[\protect\citeauthoryear{Sinha, Bai, and Ramani}{Sinha
  et~al\mbox{.}}{2016}]%
        {sinha2016deep}
\bibfield{author}{\bibinfo{person}{Ayan Sinha}, \bibinfo{person}{Jing Bai},
  {and} \bibinfo{person}{Karthik Ramani}.} \bibinfo{year}{2016}\natexlab{}.
\newblock \showarticletitle{Deep learning 3D shape surfaces using geometry
  images}. In \bibinfo{booktitle}{\emph{European Conference on Computer
  Vision}}. Springer, \bibinfo{pages}{223--240}.
\newblock


\bibitem[\protect\citeauthoryear{Su, Maji, Kalogerakis, and Learned-Millers}{Su
  et~al\mbox{.}}{2015}]%
        {Su15Multi}
\bibfield{author}{\bibinfo{person}{Hang Su}, \bibinfo{person}{Subhransu Maji},
  \bibinfo{person}{Evangelos Kalogerakis}, {and} \bibinfo{person}{Erik
  Learned-Millers}.} \bibinfo{year}{2015}\natexlab{}.
\newblock \showarticletitle{Multi-view Convolutional Neural Networks for 3D
  Shape Recognition}. In \bibinfo{booktitle}{\emph{International Conference on
  Computer Vision (ICCV)}}.
\newblock


\bibitem[\protect\citeauthoryear{Such, Sah, Dominguez, Pillai, Zhang, Michael,
  Cahill, and Ptucha}{Such et~al\mbox{.}}{2017}]%
        {Such17Robust}
\bibfield{author}{\bibinfo{person}{F.~P. Such}, \bibinfo{person}{S. Sah},
  \bibinfo{person}{M.~A. Dominguez}, \bibinfo{person}{S. Pillai},
  \bibinfo{person}{C. Zhang}, \bibinfo{person}{A. Michael},
  \bibinfo{person}{N.~D. Cahill}, {and} \bibinfo{person}{R. Ptucha}.}
  \bibinfo{year}{2017}\natexlab{}.
\newblock \showarticletitle{Robust Spatial Filtering With Graph Convolutional
  Neural Networks}.
\newblock \bibinfo{journal}{\emph{IEEE Journal of Selected Topics in Signal
  Processing}} \bibinfo{volume}{11}, \bibinfo{number}{6} (\bibinfo{date}{Sept}
  \bibinfo{year}{2017}), \bibinfo{pages}{884--896}.
\newblock


\bibitem[\protect\citeauthoryear{Tarini, Pietroni, Cignoni, Panozzo, and
  Puppo}{Tarini et~al\mbox{.}}{2010}]%
        {tarini2010practical}
\bibfield{author}{\bibinfo{person}{Marco Tarini}, \bibinfo{person}{Nico
  Pietroni}, \bibinfo{person}{Paolo Cignoni}, \bibinfo{person}{Daniele
  Panozzo}, {and} \bibinfo{person}{Enrico Puppo}.}
  \bibinfo{year}{2010}\natexlab{}.
\newblock \showarticletitle{Practical quad mesh simplification}. In
  \bibinfo{booktitle}{\emph{Computer Graphics Forum}},
  Vol.~\bibinfo{volume}{29}. Wiley Online Library, \bibinfo{pages}{407--418}.
\newblock


\bibitem[\protect\citeauthoryear{Tatarchenko, Park, Koltun, and
  Zhou}{Tatarchenko et~al\mbox{.}}{2018}]%
        {Tat18}
\bibfield{author}{\bibinfo{person}{Maxim Tatarchenko}, \bibinfo{person}{Jaesik
  Park}, \bibinfo{person}{Vladlen Koltun}, {and} \bibinfo{person}{Qian-Yi
  Zhou}.} \bibinfo{year}{2018}\natexlab{}.
\newblock \showarticletitle{Tangent Convolutions for Dense Prediction in 3D}.
  In \bibinfo{booktitle}{\emph{Proceedings of the IEEE Conference on Computer
  Vision and Pattern Recognition}}. \bibinfo{pages}{3887--3896}.
\newblock


\bibitem[\protect\citeauthoryear{Tchapmi, Choy, Armeni, Gwak, and
  Savarese}{Tchapmi et~al\mbox{.}}{2017}]%
        {Tchapmi17SEGCloud}
\bibfield{author}{\bibinfo{person}{Lyne~P. Tchapmi},
  \bibinfo{person}{Christopher~B. Choy}, \bibinfo{person}{Iro Armeni},
  \bibinfo{person}{JunYoung Gwak}, {and} \bibinfo{person}{Silvio Savarese}.}
  \bibinfo{year}{2017}\natexlab{}.
\newblock \showarticletitle{SEGCloud: Semantic Segmentation of 3D Point
  Clouds}. In \bibinfo{booktitle}{\emph{3DV}}.
\newblock


\bibitem[\protect\citeauthoryear{Velickovic, Cucurull, Casanova, Romero, Lio,
  and Bengio}{Velickovic et~al\mbox{.}}{2018}]%
        {Velickovic2018graph}
\bibfield{author}{\bibinfo{person}{Petar Velickovic}, \bibinfo{person}{Guillem
  Cucurull}, \bibinfo{person}{Arantxa Casanova}, \bibinfo{person}{Adriana
  Romero}, \bibinfo{person}{Pietro Lio}, {and} \bibinfo{person}{Yoshua
  Bengio}.} \bibinfo{year}{2018}\natexlab{}.
\newblock \showarticletitle{Graph Attention Networks}. In
  \bibinfo{booktitle}{\emph{International Conference on Learning
  Representations}}.
\newblock


\bibitem[\protect\citeauthoryear{Verma, Boyer, and Verbeek}{Verma
  et~al\mbox{.}}{2018}]%
        {Verma2018FeaStNetFG}
\bibfield{author}{\bibinfo{person}{Nitika Verma}, \bibinfo{person}{E. Boyer},
  {and} \bibinfo{person}{Jakob Verbeek}.} \bibinfo{year}{2018}\natexlab{}.
\newblock \showarticletitle{FeaStNet: Feature-Steered Graph Convolutions for 3D
  Shape Analysis}. In \bibinfo{booktitle}{\emph{CVPR}}.
\newblock


\bibitem[\protect\citeauthoryear{Vlasic, Baran, Matusik, and
  Popovi{\'c}}{Vlasic et~al\mbox{.}}{2008}]%
        {vlasic2008articulated}
\bibfield{author}{\bibinfo{person}{Daniel Vlasic}, \bibinfo{person}{Ilya
  Baran}, \bibinfo{person}{Wojciech Matusik}, {and} \bibinfo{person}{Jovan
  Popovi{\'c}}.} \bibinfo{year}{2008}\natexlab{}.
\newblock \showarticletitle{Articulated mesh animation from multi-view
  silhouettes}. In \bibinfo{booktitle}{\emph{ACM Transactions on Graphics
  (TOG)}}, Vol.~\bibinfo{volume}{27}. ACM, \bibinfo{pages}{97}.
\newblock


\bibitem[\protect\citeauthoryear{Wang, Liu, Guo, Sun, and Tong}{Wang
  et~al\mbox{.}}{2017}]%
        {Wang17OCNN}
\bibfield{author}{\bibinfo{person}{Peng-Shuai Wang}, \bibinfo{person}{Yang
  Liu}, \bibinfo{person}{Yu-Xiao Guo}, \bibinfo{person}{Chun-Yu Sun}, {and}
  \bibinfo{person}{Xin Tong}.} \bibinfo{year}{2017}\natexlab{}.
\newblock \showarticletitle{O-CNN: Octree-based Convolutional Neural Networks
  for 3D Shape Analysis}.
\newblock \bibinfo{journal}{\emph{ACM Trans. Graph.}} \bibinfo{volume}{36},
  \bibinfo{number}{4}, Article \bibinfo{articleno}{72} (\bibinfo{date}{July}
  \bibinfo{year}{2017}), \bibinfo{numpages}{11}~pages.
\newblock
\showISSN{0730-0301}
\urldef\tempurl%
\url{https://doi.org/10.1145/3072959.3073608}
\showDOI{\tempurl}


\bibitem[\protect\citeauthoryear{Wang, Asafi, van Kaick, Zhang, Cohen-Or, and
  Chen}{Wang et~al\mbox{.}}{2012}]%
        {wang2012active}
\bibfield{author}{\bibinfo{person}{Yunhai Wang}, \bibinfo{person}{Shmulik
  Asafi}, \bibinfo{person}{Oliver van Kaick}, \bibinfo{person}{Hao Zhang},
  \bibinfo{person}{Daniel Cohen-Or}, {and} \bibinfo{person}{Baoquan Chen}.}
  \bibinfo{year}{2012}\natexlab{}.
\newblock \showarticletitle{Active co-analysis of a set of shapes}.
\newblock \bibinfo{journal}{\emph{ACM Transactions on Graphics (TOG)}}
  \bibinfo{volume}{31}, \bibinfo{number}{6} (\bibinfo{year}{2012}),
  \bibinfo{pages}{165}.
\newblock


\bibitem[\protect\citeauthoryear{Wang, Sun, Liu, Sarma, Bronstein, and
  Solomon}{Wang et~al\mbox{.}}{2018a}]%
        {dgcnn}
\bibfield{author}{\bibinfo{person}{Yue Wang}, \bibinfo{person}{Yongbin Sun},
  \bibinfo{person}{Ziwei Liu}, \bibinfo{person}{Sanjay~E Sarma},
  \bibinfo{person}{Michael~M Bronstein}, {and} \bibinfo{person}{Justin~M
  Solomon}.} \bibinfo{year}{2018}\natexlab{a}.
\newblock \showarticletitle{Dynamic graph CNN for learning on point clouds}.
\newblock \bibinfo{journal}{\emph{arXiv preprint arXiv:1801.07829}}
  (\bibinfo{year}{2018}).
\newblock


\bibitem[\protect\citeauthoryear{Wang, Sun, Liu, Sarma, Bronstein, and
  Solomon}{Wang et~al\mbox{.}}{2018b}]%
        {wang2018dynamic}
\bibfield{author}{\bibinfo{person}{Yue Wang}, \bibinfo{person}{Yongbin Sun},
  \bibinfo{person}{Ziwei Liu}, \bibinfo{person}{Sanjay~E Sarma},
  \bibinfo{person}{Michael~M Bronstein}, {and} \bibinfo{person}{Justin~M
  Solomon}.} \bibinfo{year}{2018}\natexlab{b}.
\newblock \showarticletitle{Dynamic Graph CNN for Learning on Point Clouds}.
\newblock \bibinfo{journal}{\emph{arXiv preprint arXiv:1801.07829}}
  (\bibinfo{year}{2018}).
\newblock


\bibitem[\protect\citeauthoryear{Williams, Schneider, Silva, Zorin, Bruna, and
  Panozzo}{Williams et~al\mbox{.}}{2018}]%
        {williams2018deep}
\bibfield{author}{\bibinfo{person}{Francis Williams}, \bibinfo{person}{Teseo
  Schneider}, \bibinfo{person}{Claudio Silva}, \bibinfo{person}{Denis Zorin},
  \bibinfo{person}{Joan Bruna}, {and} \bibinfo{person}{Daniele Panozzo}.}
  \bibinfo{year}{2018}\natexlab{}.
\newblock \showarticletitle{Deep Geometric Prior for Surface Reconstruction}.
\newblock \bibinfo{journal}{\emph{arXiv preprint arXiv:1811.10943}}
  (\bibinfo{year}{2018}).
\newblock


\bibitem[\protect\citeauthoryear{Wu, Song, Khosla, Yu, Zhang, Tang, and
  Xiao}{Wu et~al\mbox{.}}{2015}]%
        {Wu15Shapenet}
\bibfield{author}{\bibinfo{person}{Zhirong Wu}, \bibinfo{person}{Shuran Song},
  \bibinfo{person}{Aditya Khosla}, \bibinfo{person}{Fisher Yu},
  \bibinfo{person}{Linguang Zhang}, \bibinfo{person}{Xiaoou Tang}, {and}
  \bibinfo{person}{Jianxiong Xiao}.} \bibinfo{year}{2015}\natexlab{}.
\newblock \showarticletitle{{3D} shapenets: A deep representation for
  volumetric shapes}. In \bibinfo{booktitle}{\emph{Computer Vision and Pattern
  Recognition (CVPR)}}. \bibinfo{pages}{1912–1920}.
\newblock


\bibitem[\protect\citeauthoryear{Xu, Dong, and Zhong}{Xu et~al\mbox{.}}{2017}]%
        {xu2017directionally}
\bibfield{author}{\bibinfo{person}{Haotian Xu}, \bibinfo{person}{Ming Dong},
  {and} \bibinfo{person}{Zichun Zhong}.} \bibinfo{year}{2017}\natexlab{}.
\newblock \showarticletitle{Directionally Convolutional Networks for 3D Shape
  Segmentation}. In \bibinfo{booktitle}{\emph{Proceedings of the IEEE
  International Conference on Computer Vision}}. \bibinfo{pages}{2698--2707}.
\newblock


\bibitem[\protect\citeauthoryear{Yi, Su, Guo, and Guibas}{Yi
  et~al\mbox{.}}{2017}]%
        {Yi17SyncSpecCNN}
\bibfield{author}{\bibinfo{person}{Li Yi}, \bibinfo{person}{Hao Su},
  \bibinfo{person}{Xingwen Guo}, {and} \bibinfo{person}{Leonidas Guibas}.}
  \bibinfo{year}{2017}\natexlab{}.
\newblock \showarticletitle{SyncSpecCNN: Synchronized Spectral CNN for 3D Shape
  Segmentation}. In \bibinfo{booktitle}{\emph{Computer Vision and Pattern
  Recognition (CVPR)}}.
\newblock


\bibitem[\protect\citeauthoryear{Ying, You, Morris, Ren, Hamilton, and
  Leskovec}{Ying et~al\mbox{.}}{2018}]%
        {Ying18Hierarchical}
\bibfield{author}{\bibinfo{person}{Zhitao Ying}, \bibinfo{person}{Jiaxuan You},
  \bibinfo{person}{Christopher Morris}, \bibinfo{person}{Xiang Ren},
  \bibinfo{person}{Will Hamilton}, {and} \bibinfo{person}{Jure Leskovec}.}
  \bibinfo{year}{2018}\natexlab{}.
\newblock \showarticletitle{Hierarchical Graph Representation Learning with
  Differentiable Pooling}. In \bibinfo{booktitle}{\emph{Advances in Neural
  Information Processing Systems}}. \bibinfo{pages}{4805--4815}.
\newblock


\end{thebibliography}
\appendix

\section{Training Configurations}
\label{sec:meshcnncfg}
\rg{For classification we use the same network architecture for the SHREC and Cube engraving datasets. 
We detail the network configurations and learning parameters in Table~\ref{tab:cfg}. 
For the segmentation task, for both the COSEG and human body datasets, we use a Unet~\cite{ronneberger2015u} type network. Table~\ref{tab:seg_cfg} provides the details of this network.
}
\begin{table}
\begin{center}
\begin{tabular}{c}
\begin{tabular}{c} 
\hline\hline
MeshConv $f_{in} \times 32$\\ 
MeshPool $\xrightarrow{} 600$ \\ 
MeshConv $32 \times 64$ \\ 
MeshPool $\xrightarrow{} 450$ \\ 
MeshConv $64 \times 128$ \\ 
MeshPool $\xrightarrow{} 300$ \\ 
MeshConv $128 \times 256$ \\ 
MeshPool $\xrightarrow{} 279$ \\ 
GlobalAvgPool \\
FC $ 256  \times 100 $ \\ 
FC $ 100 \times 30 $ \\ 
\hline\hline
Classification
\end{tabular} 
\end{tabular} 
\end{center}
\caption{Classification network configuration for SHREC and Cube engraving. Networks start with $750$ edges. They both use Adam optimization, $lr=0.0002$, and group norm ($g=16$). The data augmentation is with $5 \%$ edge flips and $20\%$ slide vertices.}
\label{tab:cfg}
\end{table}

\begin{table}
\begin{center}
\begin{tabular}{c}
\begin{tabular}{c} 
\hline\hline
ResConv $f_{in} \times 32$\\ 
MeshPool $\xrightarrow{} 1200$ \\ 
ResConv $32 \times 64$ \\ 
MeshPool $\xrightarrow{} 900$ \\ 
ResConv $64 \times 128$ \\ 
MeshPool $\xrightarrow{} 300$ \\ 
ResConv $128 \times 256$ \\ 
MeshPool $\xrightarrow{} 279$ \\ 
\hline\hline
Segmentation
\end{tabular}


\end{tabular} 
\end{center}
\caption{Segmentation network configuration for COSEG and human body datasets. Networks start with $2250$ edges. Only down part in the network is presented. The up part is symmetric to the down part. 
}
\label{tab:seg_cfg}
\end{table}




\end{document}